%% file: 0_Main.tex
\documentclass[journal ]{new-aiaa}
\usepackage[utf8]{inputenc}
\usepackage{textcomp}

\usepackage{graphicx}
\usepackage{amsmath}
\usepackage[version=4]{mhchem}
\usepackage{siunitx}
\usepackage{longtable,tabularx}
\usepackage[acronym]{glossaries}

\usepackage{bm,mathtools}

\usepackage{color}

\usepackage{caption}
\usepackage{subcaption}

\usepackage{booktabs}
\usepackage{multirow}

\usepackage{comment}

\usepackage[linesnumbered,ruled,vlined]{algorithm2e}
\usepackage{cases}
\usepackage{hyperref}
\usepackage[capitalise]{cleveref}
\usepackage[export]{adjustbox}
\usepackage{xcolor}

\usepackage[printonlyused]{acronym}

\usepackage{dirtytalk}
\usepackage{soul}
\newcommand{\subsubsubsection}[1]{\paragraph{#1}\mbox{}\\}

\title{Treatment of Epistemic Uncertainty in Conjunction Analysis with Dempster-Shafer Theory}

\author{L. Sanchez \footnote{PhD Candidate, Aerospace Centre of Excellence, luis.sanchez-fdez-mellado@strath.ac.uk.} and M. Vasile\footnote{Professor at University of Strathclyde and Director of Aerospace Centre of Excellence, massimiliano.vasile@strath.ac.uk.}}
\affil{Aerospace Centre of Excellence. University of Strathclyde, Glasgow, G1 1XJ, United Kingdom}
\author{S. Sanvido\footnote{Space Debris Engineer. IMS Space Consultancy GmbH, silvia.sanvido@ext.esa.int.}}
\affil{Space IMS Space Consultancy GmbH, 64297, Darmstadt, Germany}
\author{K. Mertz \footnote{Spaces Debris Engineer, Space Debris Office, ESA/ESOC, klaus.merz@esa.int.}}
\affil{Space Debris Office, European Space Operations Center (ESOC), European Space Agency (ESA), Darmstadt, 64293, Germany}
\author{C. Taillan\footnote{Space Surveillance Engineer. Space Security, Safety and Sustainability Office, CNES, christophe.taillan@cnes.fr.}}
\affil{Space Security, Safety and Sustainability Office, Centre National d'Etudes Spatiales (CNES), Toulouse, 31401, France}

\begin{document}

\maketitle

\input{10_Abtract}


\input{20_Acronyms}

\input{1_introduction}

\input{2_IDSS}

\input{3_epistemicCDM}

\input{4_NumericalCases}

\input{5_conclusions}

\input{9992_acknowledgements}

\bibliography{Bibliography}

\end{document}

%% file: 10_Abtract.tex
\section*{Abstract}
\label{sec:abstract}

\begin{abstract}
    \small
    The paper presents an approach to the modelling of epistemic uncertainty in Conjunction Data Messages (CDM) and the classification of conjunction events according to the confidence in the probability of collision. The approach proposed in this paper is based on Dempster-Shafer Theory (DSt) of evidence and starts from the assumption that the observed CDMs are drawn from a family of unknown distributions. The Dvoretzky–Kiefer–Wolfowitz (DKW) inequality is used to construct robust bounds on such a family of unknown distributions starting from a time series of CDMs. A DSt structure is then derived from the probability boxes constructed with DKW inequality. The DSt structure encapsulates the uncertainty in the CDMs at every point along the time series and allows the computation of the belief and plausibility in the realisation of a given probability of collision. The methodology proposed in this paper is tested on a number of real events and compared against existing practices in the European and French Space Agencies. We will show that the classification system proposed in this paper is more conservative than the approach taken by the European Space Agency but provides an added quantification of uncertainty in the probability of collision.  
\end{abstract}

%% file: 20_Acronyms.tex
\begin{acronym}[MPC]

    \acro{AI}[AI]{Artificial Intelligence}
    
    \acro{Bel}[\textit{Bel}]{Belief}
    \acro{bpa}[\textit{bpa}]{basic probability assignment}
    
    \acro{CAM}[CAM]{Collision Avoidance Manoeuvre}
    \acro{CARA}[CARA]{Conjunction Assessment Risk Analysis}
    \acro{CDF}[CDF]{Cumulative Distribution Function}
    \acro{CDM}[CDM]{Conjunction Data Message}
    \acro{CNES}[CNES]{Centre National d'Etudes Spatiales}
    
    \acro{DKW}[DKW]{Dvoretzky–Kiefer–Wolfowitz}
    \acro{DoU}[\textit{DoU}]{Degree of Uncertainty}
    \acro{DSt}[DSt]{Dempster-Shafer theory of evidence}
    
    \acro{eCDF}[eCDF]{empirical Cumulative Distribution Function}
    \acro{ESA}[ESA]{European Space Agency}
    \acro{ESOC}[ESOC]{European Space Operations Centre}
    
    \acro{FE}[FE]{Focal Element}
    \acro{FN}[FN]{False Negative}
    \acro{FP}[FP]{False Positive}
    \acro{FPR}[FPR]{False Positive Rate}

    \acro{HBR}[HBR]{Hard-Body Radius}
    
    \acro{IDSS}[IDSS]{Intelligent Decision Support System}

    \acro{JAC}[JAC]{Java for Assessment of Conjunctions}
    
    \acro{KS}[KS]{Kolmogorov-Smirnov}

    \acro{LEO}[LEO]{Low Earth Orbit}

    \acro{ML}[ML]{Machine Learning}
    \acro{mWSM}[mWSM]{modified Weighted Sum Method}

    \acro{Pl}[\textit{Pl}]{Plausibility}
    \acro{PoC}[PoC]{Probability of Collision}

    \acro{ROC}[ROC]{Receiver Operating Characteristic}
    
    \acro{SEM}[SEM]{Space Environment Management}
    \acro{SDO}[SDO]{Space Debris Office}
    \acro{sPoC}[sPoC]{scaled Probability of Collision}
    \acro{STM}[STM]{Space Traffic Management}
    
    \acro{TCA}[TCA]{Time of Closest Approach}
    \acro{TN}[TN]{True Negative}
    \acro{TOPSIS}[TOPSIS]{Technique for Order of Preference by Similarity to Ideal Solution}
    \acro{TP}[TP]{True Positive}
    \acro{TPR}[TPR]{True Positive Rate}

    \acro{WPM}[WPM]{Weighted Product Method}
    \acro{WSM}[WSM]{Weighted Sum Method}

\end{acronym}

%% file: 1_introduction.tex
\section{Introduction}
\label{sec:introduction}

    The close encounter of two space objects, also known as a conjunction between a chaser and a target, can lead to a collision if the relative position of the two objects is not properly controlled. The \ac{PoC} to happen depends on the probability that each of the two objects occupies a given position in space. This probability can be derived from the knowledge of the orbit of the two objects and the associated uncertainty.
    
    It is customary to assume that the distribution of possible positions of the two objects at the time of closest encounter follows a multivariate Gaussian with a given mean and covariance matrix, see \cite{Merz2017,Newman2019}. This assumption is limited by three sources of uncertainty: the uncertainty in the dynamic model used to propagate the orbit from the last available observation to the time of closest approach, the uncertainty in the actual distribution at the time of closest approach, and the uncertainty in the last observed state before closest approach. We argue that all three forms of uncertainty are epistemic in nature since they derive from a lack of knowledge of the model, distribution and error in the observations.
    
    The information on a given close encounter is generally available in the form of a \ac{CDM}, which contains the means and covariances of the two objects at \ac{TCA}, see \cite{CCSDS2013}. Thus, in this paper, we start from the assumption that the mean and covariance in each \ac{CDM} are affected by epistemic uncertainty, which is reflected in an uncertainty in the correct value of the \ac{PoC}.
    
    The general attempt to compensate for the uncertainty in the \acp{CDM} is to improve the realism of the covariance matrix by improving its propagation, \cite{Aristoff2014}, or by some form of updating of the dynamic model, \cite{Cano2023}. These approaches are all very valuable but require direct access to the post-observation data. 
    Other methods based solely on the available \acp{CDM} tried to predict the next \acp{CDM} using machine learning starting from an available time series, see \cite{Pinto2020,Giacomo2021,Uriot2022,Caldas2023}, or increased the last covariance under the assumption that the series of \acp{CDM} should follow a given distribution,  \cite{Laporte2014_a,Laporte2014_b}. This last approach does not modify the mean value or miss distance.
    
    So far, only a limited number of authors have directly addressed epistemic uncertainty in conjunction analysis, see for example   \cite{Tardioli2015,Delande2018_b,Balch2019,Greco2021_a}.
    In \cite{Sanchez2020_a,Sanchez2022_b,Sanchez2022_a} the authors proposed a robust approach to conjunction analysis and collision avoidance planning based on \ac{DSt}. \ac{DSt} allows making decisions informed by the degree of confidence in the correctness of a value rather then by the value itself,  \cite{Helton2005}.
    However, the available information to build the frame of discernment that is needed in \ac{DSt} is often limited in a sequence of \acp{CDM}. \acp{CDM} contain little information on the three forms of uncertainty listed above and essentially only provide covariance and mean value of the miss distance.
    Thus, one key question is how to translate the time series of \acp{CDM} into the frame of discernment used in \ac{DSt}.
    The underlying assumption in this work is that the \acp{CDM} are observables drawn from an unknown family of distributions defined within some bounds. Without uncertainty, one would be able to exactly predict the next \acp{CDM} as the mean and covariance would only depend on observations with a known distribution and there would be no uncertainty in the propagation model and distribution at \ac{TCA}. Furthermore, we assumed that the \acp{CDM} computed from observations acquired close to the \ac{TCA} were less affected by model and distribution uncertainty. This is reasonable as the propagation time is shorter and thus both nonlinearities and model errors have a lower impact on the propagation of the distribution of the possible states.
    
    The paper introduces a methodology, based on the  \ac{DKW} inequality, \cite{Dvoretzky1956}, to derive a \ac{DSt} structure capturing the epistemic uncertainty in a given sequence of \acp{CDM}. From the \ac{DSt} structures, one can compute the \ac{Bel} and \ac{Pl} that the value of the \ac{PoC} is correct and an upper and lower bound on its value. The paper then proposes a classification system that exploits the use of \ac{Bel} and \ac{Pl} to differentiate between events that are uncertain from events that can lead to a collision.
    The overall methodology is tested on a number of real conjunction scenarios with known sequences of \acp{CDM} and compared against current practices in the \ac{ESA} and \ac{CNES}.
    
    The rest of the paper is structured as follows. \cref{sec:IDSS} briefly introduces a methodology previously presented by the authors to deal with epistemic uncertainty for risk assessment in space encounters. \cref{sec:epist_CDM} extends this methodology to deal with sequence of \acp{CDM}. In \cref{sec:num_cases}, some numerical cases are presented showing the operation of the proposed method and comparing the approach with the procedure followed by real operators. Finally, \cref{sec:conclusions} concludes the paper with the final remarks and future work.

%% file: 2_IDSS.tex
\section{Conjunction Analysis with Dempster-Shafer Structures}
\label{sec:IDSS}

    This section briefly introduces the basic idea of \ac{DSt} applied to \ac{CARA}. It also includes the \ac{DSt}-based conjunction classification system already introduced by the authors in previous works. More details on \ac{DSt} can be found in \cite{Shafer1976}, and more details on its application to space conjunction assessment can be found in \cite{Sanchez2020_a,Sanchez2022_b,Sanchez2022_a}.
    
    In this paper, we consider only fast encounters between two objects: object 1 and object 2. Under the typical modelling assumptions of fast encounters, see \cite{Serra2016}, the \ac{PoC} can be defined as: 
    \begin{equation}\label{eq:PoC}
       PoC = \frac{1}{2\pi \sqrt{\|\mathbf{\Sigma}\|}} \int\limits_{\mathcal{B}((0,0),R)} e^{ -\frac{1}{2}\left( (\mathbf{b}-\bm{\mu})^T\mathbf{\Sigma}^{-1}(\mathbf{b}-\bm{\mu}) \right)} d\xi d\zeta   
    \end{equation}
    where, without loss of generality, object 2 is at the centre of the coordinate system of the impact plane at the time of closest approach (TCA), $\mathbf{b}=[\xi,\zeta]^T$ is the position vector of object 1 with respect to object 2 projected onto the impact plane,
    $\bm\Sigma$ is the $2\times 2$ combined covariance matrix of the position of the two objects in the impact plane ($\bm\Sigma=\bm\Sigma_1+\bm\Sigma_2$, with $\bm\Sigma_1$ and $\bm\Sigma_2$ the individual covariance matrices of object 1 and 2 respectively) and $\bm\mu=[\mu_{\xi},\mu_{\zeta}]^T$ is the expected position vector of object 1 with respect to object 2 projected onto the impact plane. In the remainder of the paper $\bm\mu$ is called miss distance. The integration region ${\mathcal{B}((0,0),R)}$, or \ac{HBR}, is a disk with radius $R$ centred at the origin of the impact plane.
    
    When the covariance $\bm\Sigma$ and miss distance $\bm\mu$ are not precisely known the $PoC$ is affected by a degree of uncertainty. This lack of knowledge translates into an epistemic uncertainty in the exact value of $\bm\Sigma$ and $\bm\mu$. The epistemic uncertainty in covariance $\bm\Sigma$ and miss distance $\bm{\mu}$ can come from incertitude in the sources of information, from poor knowledge of the measurements or propagation model or from an approximation of the actual distribution on the impact plane at \ac{TCA}. As shown in \cite{Sanchez2020_a} and \cite{Sanchez2022_b}, this epistemic uncertainty can be modelled with \ac{DSt}.
    
    The idea proposed in \cite{Sanchez2020_a}, was to use \ac{DSt} to compute the level of confidence in the correctness of the value of the \ac{PoC}, given the available evidence on the sources of information. Each component of the combined covariance matrix in the impact plane, $[\sigma^2_\xi,\sigma^2_\zeta, \sigma_{\xi\zeta}]$, was modelled with one or more intervals and so was the miss distance $[\mu_\xi,\mu_\zeta]$. A \ac{bpa} was then associated with each interval.  
    The intervals and the associated \ac{bpa} can be derived, for example, directly from the raw observations, \cite{Greco2021_a,Greco2021_b}, or from a time series of \acp{CDM}, \cite{CCSDS2013}, as explained later in this paper.
    Note that in the case in which raw observation data are available, one could directly compute the confidence on the miss distance, see \cite{Greco2021_a}. However, in the following we will consider the CDMs as the observable quantities and the PoC, computed from the CDMs, to be the quantity of interest. 
    
    Given the intervals and associated \ac{bpa}, one can compute the cross-product of all the intervals under the assumption of epistemic independence. Each product of intervals with non-zero \ac{bpa} constitutes a \ac{FE}, $\gamma_i$, whose joint \ac{bpa} is the product of the \acp{bpa} of the individual intervals. 
    When computing the \ac{PoC}, each \ac{FE} defines a family of bi-variate Gaussian distributions on the impact plane. 
    In the following, the collection of all focal elements forms the uncertainty space $\textit{U}$, and the uncertain parameter vector is $\mathbf{u}=[\mu_\xi, \mu_\zeta, \sigma^2_\xi, \sigma^2_\zeta, \sigma_{\xi\zeta}]^T$ so that $\mathbf{u}\in U$.
    
    Given the set $\Phi=\{PoC|PoC\geq PoC_0\}$ and $\Omega=\{\mathbf{u}\in U|PoC(\mathbf{u})\in\Phi\}$ the \ac{Pl} and \ac{Bel} that the $PoC$ is larger than a given threshold $PoC_0$ given the available evidence are:
    \begin{subequations}\label{eq:BelPlOm}
        \begin{equation}\label{eq:BelPlOm_bel}
            Bel(\Omega) = \sum_{\gamma_i \subset \Omega}{bpa(\gamma_i)}
        \end{equation}
        \begin{equation}\label{eq:BelPlOm_pl}
            Pl(\Omega) = \sum_{\gamma_i \cap \Omega\neq \emptyset}{bpa(\gamma_i)}
        \end{equation}
    \end{subequations}
    
\begin{figure}[htb!]
        \centering
        \includegraphics[width=\textwidth]{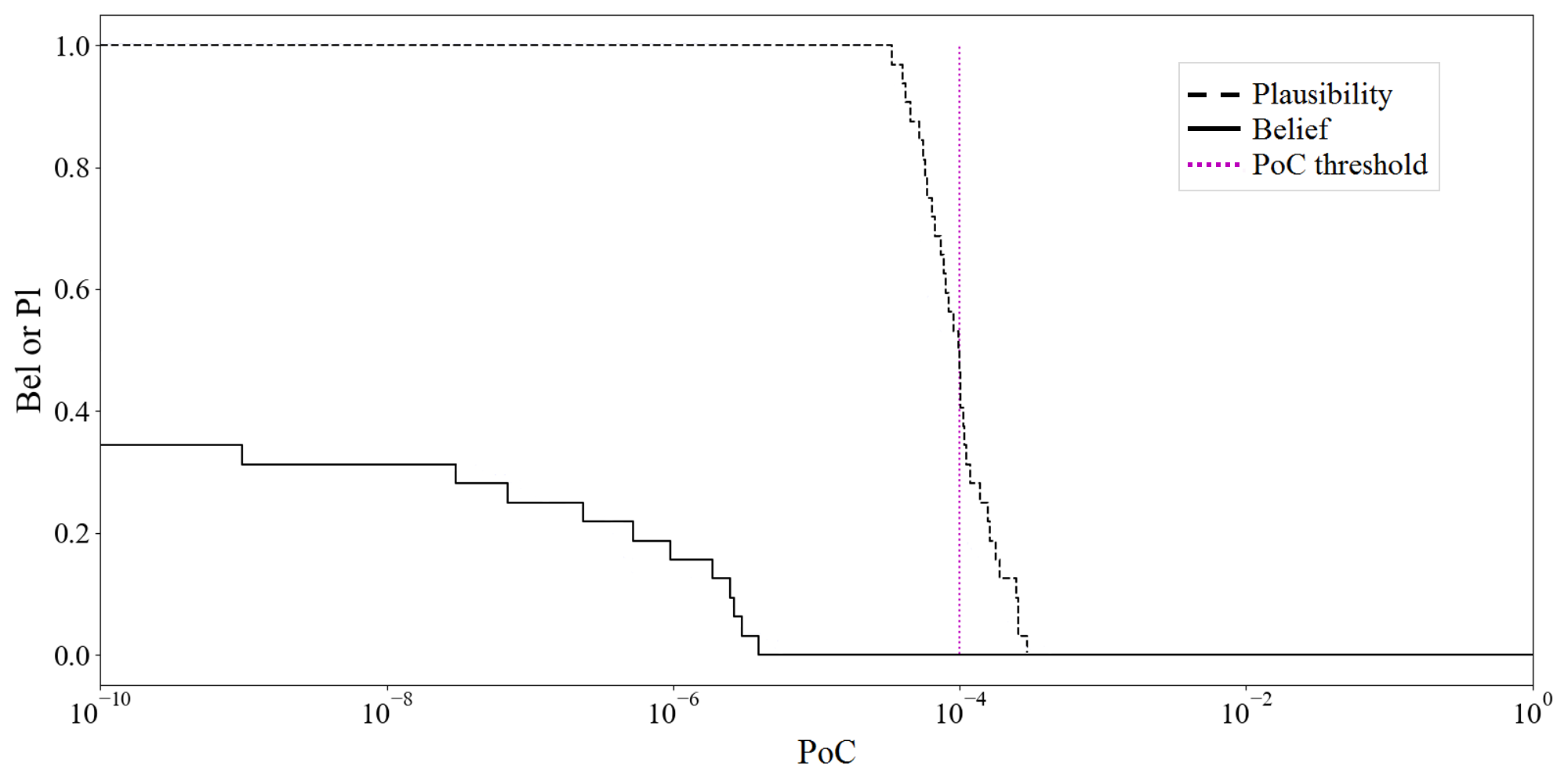}
        \captionsetup{width=0.8\linewidth,format=plain, font=small, labelfont=bf,justification=justified}
        \caption{Support to the value of $PoC$ being greater than a given value: $Bel$ -black solid line; $Pl$ - black dashed line. The dotted purple line represents a possible $PoC_0$.}
        \label{fig:PlBel_example}
    \end{figure}

    For different values $PoC_0$, \cref{eq:BelPlOm_bel,eq:BelPlOm_pl} define two curves (see the example in \cref{fig:PlBel_example}). The area between the curves, $A_{Pl,Bel}$, in logarithmic scale, is:
    \begin{equation}\label{eq:BelPlOm_area}
        A_{Pl,Bel} = \int_{\log(\underline{PoC})}^{0} Pl(\Omega) \,d(\log(PoC)) - \int_{\log(\underline{PoC})}^{0} Bel(\Omega) \,d(\log(PoC))
    \end{equation}
    
    $Bel(\Omega)$ is a lower bound on the probability that $PoC\ge PoC_0$. Its value is computed by adding up all the \acp{FE} fully supporting the hypothesis $PoC\ge PoC_0$. $Pl(\Omega)$ is an upper bound on the probability that $PoC\ge PoC_0$. Its value is computed by adding up all the \acp{FE} only partially supporting the hypothesis $PoC\ge PoC_0$. 
    The area $A_{Pl,Bel}$ quantifies the amount of uncertainty on the probability that $PoC\ge PoC_0$, i.e. if no epistemic uncertainty is present, both curves would reduced to the same \ac{CDF}. Thus, for a given value of $PoC_0$, a large value of \ac{Pl} associated with a small value of $A_{Pl,Bel}$ suggests that there is a lot of support to the hypothesis $PoC\ge PoC_0$ given the available information. On the contrary a large value of \ac{Pl} associated to a large value of $A_{Pl,Bel}$ suggests that the hypothesis $PoC\ge PoC_0$ is very plausible to be true but with a high degree of uncertainty. 
    
    In \cite{Sanchez2020_a}, a \ac{DSt}-based classification system was proposed to decide whether, for a given conjunction event, a \ac{CAM} was required or not. In this paper, we propose a revised version of the classification approach proposed \cite{Sanchez2020_a}. 
    A given conjunction event is classified according to: i) the value of the \ac{Pl} at $PoC=PoC_0$ or $Pl(PoC_0)$, ii) the time of closest approach $t2TCA$ and iii) the area $A_{Pl,Bel}$. We introduced five thresholds: two time thresholds indicating the proximity of the event, $T_1$ and $T_2$, the maximum admissible $PoC$, or $PoC_0$, the level of \ac{Pl}, $Pl_0$, above which there is sufficient support to the hypothesis $PoC\ge PoC_0$, and the value of area $A_0$, above which the information is considered to be uncertain. Three of the five thresholds, $T_1$, $T_2$ and $PoC_0$, are decided by the operators and depend on operational constraints, the other two $Pl_0$ and $A_0$ need to be tuned under evidence-based criteria, as it will be explained in the remainder of the paper. 
    
    We then introduce the following six classes, see \cref{table:Crit}, each defined by a combination of $Pl(PoC_0)$, $t2TCA$ and $A_{Pl,Bel}$: 
    \begin{table}[hbt!]
        \begin{center}
            %\small
            \captionsetup{width=0.8\linewidth,format=plain, font=small, labelfont=bf, justification=justified}
            \caption{Conjunction risk assessment evidence-based classification criterion.}
            \begin{tabular}{ l l l l l }
                \toprule
                    \textbf{Time to TCA} & $\mathbf{Pl}$ \textbf{at} $\mathbf{PoC_0}$ & \textbf{Area between curves} &   \textbf{Class}\\
                \midrule
                    \multirow{3}{*}{$t2TCA \leq T_1$} & \multirow{1}{*}{$Pl(PoC_0) < Pl_0$} &                                    \multicolumn{1}{c}{-} & 5 \\ 
                    & \multirow{2}{*}{$Pl(PoC_0) \geq Pl_0$}    & $A_{Pl,Bel}  < A_0$ & 1 \\ 
                                        & & $A_{Pl,Bel}  \geq A_0$    & 0 \\
                    \\
                    \multirow{3}{*}{\vtop{\hbox{\strut $T_1 < t2TCA$} \hbox{\strut $t2TCA \leq T_2$}}} & \multirow{1}{*}{$Pl(PoC_0) < Pl_0$} &                                  \multicolumn{1}{c}{-} & 4 \\ 
                    & \multirow{2}{*}{$Pl(PoC_0) \geq Pl_0$}    & $A_{Pl,Bel}  < A_0$ & 2 \\ 
                                        & & $A_{Pl,Bel}  \geq A_0$    & 3 \\
                    \\
                   $t2TCA > T_2$ &  \multicolumn{1}{c}{-} &  \multicolumn{1}{c}{-} & 3 \\ 
                \bottomrule
            \end{tabular}
            \label{table:Crit}
        \end{center}
    \end{table}
    
    \begin{itemize}
        \item \textit{Class 0}: there is enough evidence supporting $PoC\ge PoC_0$ but is accompanied by a high degree of uncertainty and no time to acquire new measurements, due to the proximity of the event, hence a \ac{CAM} is recommended.
        \item \textit{Class 1}: there is full support to the hypothesis $PoC\ge PoC_0$, with limited uncertainty, and short $t2TCA$, hence a \ac{CAM} is required.
        \item \textit{Class 2}: there is full support to the hypothesis $PoC\ge PoC_0$, with limited uncertainty, preparing a \ac{CAM} is recommended, but a \ac{CAM} is not executed yet due to the available time before the encounter.
        \item \textit{Class 3}: there is enough evidence supporting $PoC\ge PoC_0$ but is accompanied by a high degree of uncertainty with sufficient time to acquire new measurements, hence more measurements should be acquired.
        \item \textit{Class 4}: there is insufficient evidence supporting $PoC\ge PoC_0$ and sufficient time to acquire new measurements.
        \item \textit{Class 5}: no action is implemented, since $t2TCA$ is too short and there is insufficient evidence supporting $PoC\ge PoC_0$.
    \end{itemize}
    Note that for $t2TCA>T_2$ all events are classified as \textit{Class 3} because the required action is to acquire more measurements. Also, it has to be noted that the level of confidence that one has in the computed value of the $PoC$ depends only on $Pl_0$. If $Pl_0$ is set to zero it means that one accepts even a single piece of partial evidence that $PoC\ge PoC_0$ to escalate the \textit{Class} from 5, to 0 or 1, or from 4 to 2 or 3.

%% file: 3_epistemicCDM.tex
\section{Modelling Epistemic Uncertainty in Conjunction Data Messages}
\label{sec:epist_CDM}

    The use of \ac{DSt} to model epistemic uncertainty does not require any assumption on the probability of an event and also captures rare events with low probability. On the other hand with no direct information on measurements and dynamic model, one can only rely on the \acp{CDM} to define the \acp{FE} and associated probability masses.
    
    This section presents a methodology to associate one of the six classes introduced in the previous section to a given sequence of \acp{CDM}.
    The first step is to derive the \acp{FE} from the time series of miss distances and covariance matrices in the \acp{CDM}. In accordance with \ac{DSt}, we make no prior assumption on the underlying distribution of the \acp{CDM} and, instead, we consider that each \ac{CDM} is drawn from an unknown set of probability distributions. The assumption is that the value of the uncertain vector $\mathbf{u}$ in each \acp{CDM} is a sample drawn from the set of unknown distributions. 
    We make use of the \ac{DKW} inequality, \cite{Dvoretzky1956}, to build an upper and lower bound to the set starting from the \ac{eCDF} derived from the sequence of \acp{CDM}. 
    
\begin{figure}[htb!]
        \centering
        \begin{subfigure}[t]{0.49\textwidth}
            \centering
            \includegraphics[width=\textwidth]{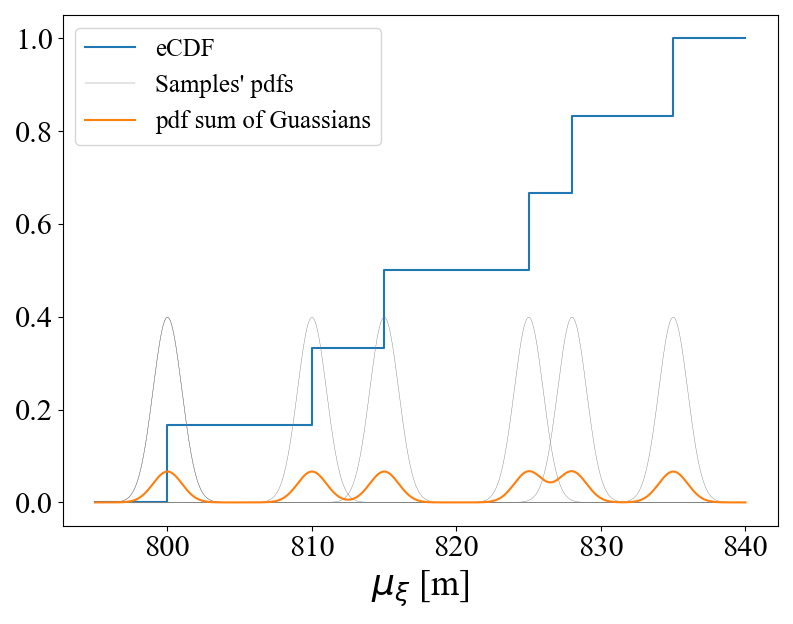}
            \caption{}
            \label{fig:eCDF_a}
        \end{subfigure}
        \begin{subfigure}[t]{0.49\textwidth}
            \centering
            \includegraphics[width=\textwidth]{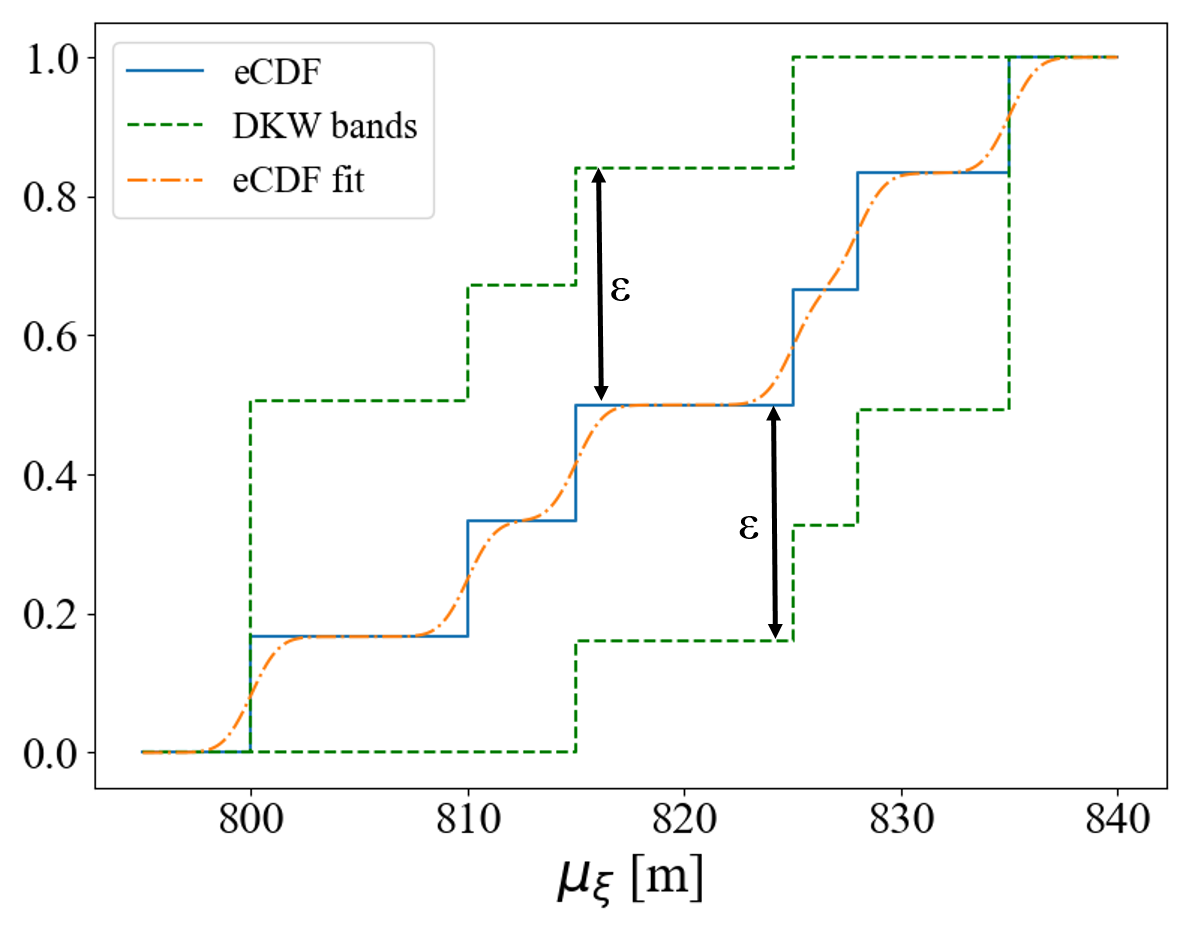}
            \caption{}
            \label{fig:eCDF_b}
        \end{subfigure}
        \newline
        \begin{subfigure}[t]{0.49\textwidth}
            \centering
            \includegraphics[width=\textwidth]{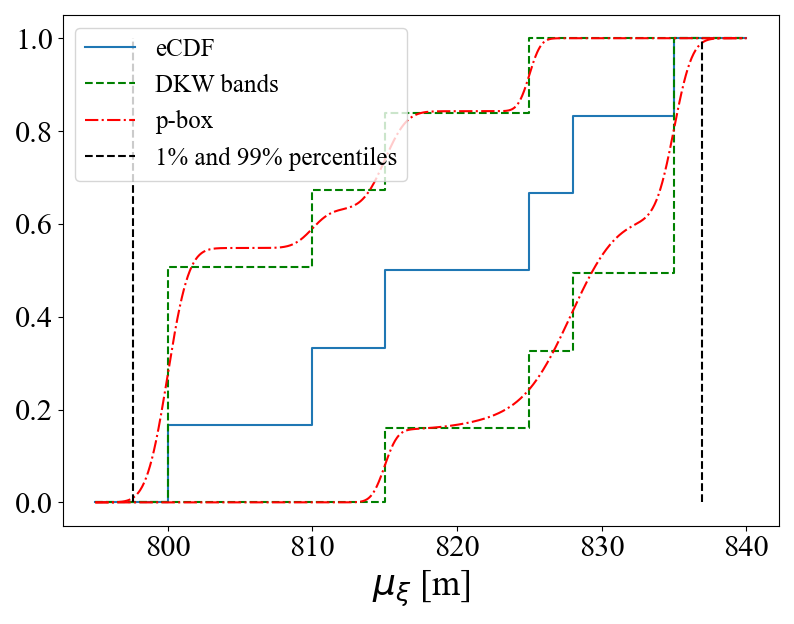}
            \caption{}
            \label{fig:eCDF_c}
        \end{subfigure}
        \begin{subfigure}[t]{0.49\textwidth}
            \centering
            \includegraphics[width=\textwidth]{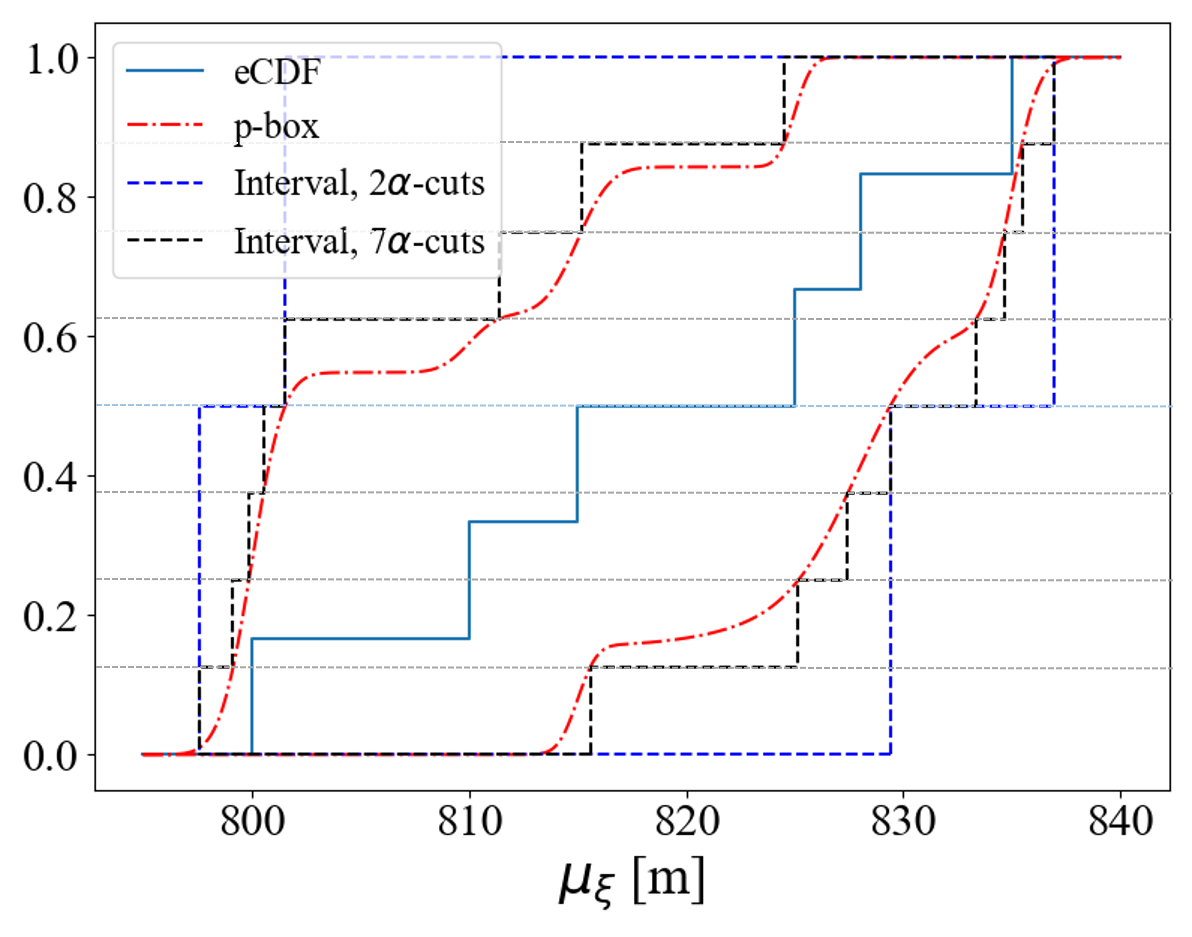}
            \caption{}
            \label{fig:eCDF_d}
        \end{subfigure}
        \captionsetup{width=0.8\linewidth,format=plain, font=small, labelfont=bf,justification=justified}
        \caption{Example of intervals derivation form the eCDF. (a) eCDF (solid blue), individual sample's Gaussian pdf distributions (solid grey), pdf of the sum of Gaussian distributions for the eCDF fit (solid orange) (b) eCDF (solid blue), DKW bands (dashed green), fitted eCDF with weighted sum of Gaussian distributions (dashed-pointed orange). (c) eCDF (solid blue), DKW bands (dashed green), p-box optimising the weighted sum of Gaussian distributions (dashed-pointed red), 1\% and 99\% percentiles (vertical pointed black lines). (d) eCDF (solid blue), p-box (dashed-pointed red), 1 $\alpha$-cut 2 intervals' Pl and Bel (dashed blue), 7 $\alpha$-cuts 8 intervals' Pl and Bel (dashed black). Dotted thin horizontal lines for the $\alpha$-cuts: light blue at 0.5 for the 2 intervals partition, grey lines spaced 0.125 for the 8 intervals partition.}
        \label{fig:eCDF}
    \end{figure}

    Given a sequence of \acp{CDM} and the \ac{eCDF} of each of the components of the uncertain vector $\mathbf{u}$,
    the \ac{DKW} inequality defines the following upper and a lower bounds 
    \begin{equation}\label{eq:DWT_band}
        F_n(x) - \sqrt{\frac{\ln{\frac{2}{\delta}}}{2n}} \leq \mathcal{F}(x) \leq F_n(x)+\sqrt{\frac{\ln{\frac{2}{\delta}}}{2n}}
    \end{equation}
    around the \ac{eCDF} $F_n(x)$
    (dashed green lines in \cref{fig:eCDF_b}), given $n$ \acp{CDM} and the confidence level $1-\delta$ that the exact distribution $\mathcal{F}(x) \in F_n(x) \pm \varepsilon$, where $\varepsilon=\sqrt{\frac{\ln{\frac{2}{\delta}}}{2n}}$.
    
    Note that expression \cref{eq:DWT_band} implies that for an infinite number of observations $\mathcal{F}(x)= F_n(x)$. However, in the following, we will show that in real sequences not all \acp{CDM} follow the same distribution. Convergence to a single distribution is, therefore, plausible for a single sequence with consistent measurements and propagation model. Furthermore, $F_n(x)$ would converge to a delta function if each observation returned the same mean and covariance and the propagation model would not introduce any variability or nonlinearity. 

    From the confidence region defined by the \ac{DKW} bands, it is possible to build a probability box, or p-box,  \cite{Ferson2003,Ferson2007,Liu2017}, for each of the components of $\mathbf{u}$. A p-box is a set of all \acp{CDF} compatible with the data, that is, the bounded region containing all distributions from where the set of samples may have been drawn,  \cite{Ferson2007}.  The upper and lower bounds of the p-box are monotonic non-decreasing functions, ranging from 0 and 1, so that $\underline{\mathcal{F}}(x)\leq\mathcal{F}(x)\leq\overline{\mathcal{F}}(x)$, with $\underline{\mathcal{F}}(x)$ and $\overline{\mathcal{F}}(x)$ the upper and lower bounds of the p-box for a given variable $x$,  \cite{Ferson2003}. 
    
    In this work, the p-box bounds are computed from the \ac{CDF} of a weighted sum of univariate Gaussians, each one centred at one of the samples. More formally the assumption is that $\mathcal{F}(x)$ can be approximated by:
    \begin{equation}\label{eq:weighted_sum_Gauss}
        \mathcal{F}(x)\sim\mathcal{P}(x) = \int_{-\inf}^{\inf} \sum_{i}^n w_i\mathcal{N}(x_i,\sigma_i;x) \,dx ,
    \end{equation}
    with $x_i$ the realisations of the uncertain variable $x$, $w_i$ a weight associated with each sample, and $\sigma_i$ the variance of the Gaussian distribution associated with the ith-sample. See \cref{fig:eCDF_a} for an illustrative example. 
    Implicitly, it implies that each sample presents some uncertainty which is modelled with a Gaussian distribution (grey lines in \cref{fig:eCDF_a}). This distribution represents the confidence in the sample's value. 
    By doing so, we admit that when we observe a sequence of \ac{CDM} we cannot tell from which exact distribution that sequence is drawn. This is consistent with the available sequences of real \acp{CDM} and the approach adopted by \ac{CNES} to model the uncertainty in the covariance realism (see section \ref{sec:CNES_example}).
    
    In order to define the limits of the p-box, the two free parameters on each Gaussian distribution on the weighted sum, $w_i$ and $\sigma_i$, must be computed by solving the optimisation problems:
    \begin{equation}\label{eq:p-box_bounds}
        \left\{\begin{array}{l}
             \overline{\mathcal{P}}(x) = \max_{w_i,\sigma_i} \mathcal{P}(x;w_i,\sigma_i)  \\
             \underline{\mathcal{P}}(x)  = \min_{w_i,\sigma_i} \mathcal{P}(x;w_i,\sigma_i)
        \end{array}\right.
        s.t. 
        \left\{\begin{array}{l}
             \overline{\mathcal{P}}(x) \leq \min(1,F_n(x)+\varepsilon)  \\
             \underline{\mathcal{P}}(x) \geq \max(0,F_n(x)-\epsilon)         
        \end{array}\right. \text{,}
    \end{equation}
    where $\overline{\mathcal{P}}(x), \underline{\mathcal{P}}(x)$ are the upper and lower bounds of the p-box, respectively (red dashed-pointed line in \cref{fig:eCDF_c}).
    An approximation to $\overline{\mathcal{P}}(x), \underline{\mathcal{P}}(x)$ can be computed by finding the values of $w_i$ and $\sigma_i$ in \cref{eq:weighted_sum_Gauss} that best fit the upper and lower \ac{DKW} bands:
    \begin{equation}\label{eq:approx_p-box_bounds}
        \left\{ \begin{array}{l}
             \overline{\mathcal{P}}(x)\approx\overline{P}(x) = \text{fit}_{w_i,\sigma_i} (F_n(x)+\epsilon) \\
             \underline{\mathcal{P}}(x)\approx\underline{P}(x) = \text{fit}_{w_i,\sigma_i} (F_n(x)-\epsilon)
        \end{array}\right. \text{.}
    \end{equation}
    \cref{eq:approx_p-box_bounds} gives the upper and lower bounds on the probability of realising a particular value of the uncertain vector $\mathbf{u}$ but the definition of a set of intervals for each component of $\mathbf{u}$ requires first the definition of the range of each component.
    \cref{eq:weighted_sum_Gauss} suggests that each p-box has infinite support. However, this would lead to an inconvenient infinite range for variance and miss distance. Instead, in the following we define the more practical interval $[\underline{x},\overline{x}]$ such that:
    \begin{equation}\label{eq:weighted_sum_Gauss_left}
         \int_{\underline{x}}^{\infty} w_1\mathcal{N}(x_1,\sigma_1;x) \,dx = 0.99,\;\; \int_{-\infty}^{\overline{x}} w_n\mathcal{N}(x_n,\sigma_i;x) \,dx = 0.99.
    \end{equation}
    
    It is important to note that the assumption is that the miss distance and each component of the covariance can be treated independently. This is generally not the case, however, the independence assumption in this paper leads to a more conservative set of focal elements that cover the space of realisations of the uncertainty vector. Although this can lead to over-conservative decisions, it is deemed to be acceptable in the case of high-risk events with little available information.  
    
    \subsection{Scaling of the CDMs}
\begin{figure}[htb!]
        \centering
        \begin{subfigure}[t]{0.49\textwidth}
            \centering
            \includegraphics[width=\textwidth,height=60mm]{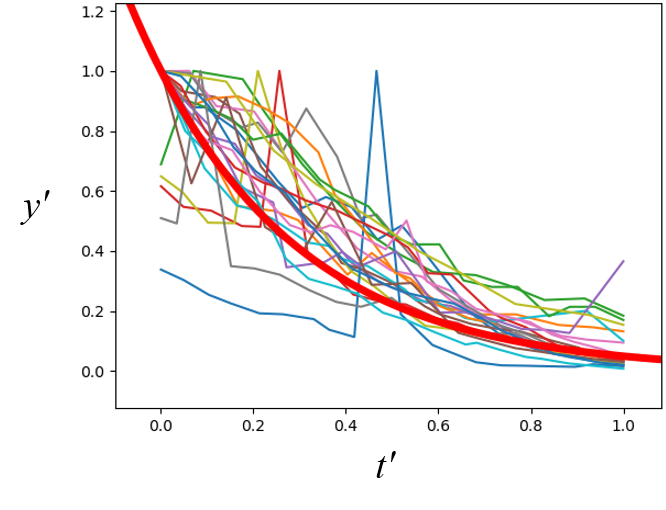}
            \caption{}
            \label{fig:fit_law_a}
        \end{subfigure}
        \begin{subfigure}[t]{0.49\textwidth}
            \centering
            \includegraphics[width=\textwidth,height=60mm]{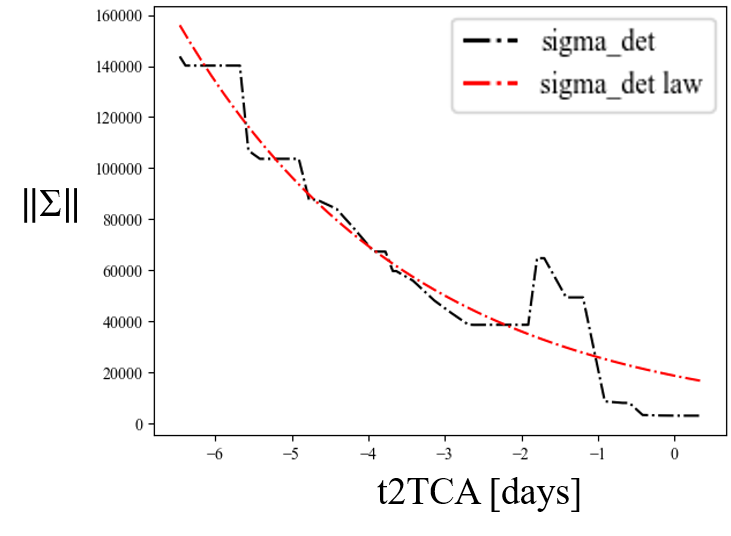}
            \caption{}
            \label{fig:fit_law_b}
        \end{subfigure}
        \captionsetup{width=0.8\linewidth,format=plain, font=small, labelfont=bf,justification=justified}
        \caption{Fitting law: (a) $y'=e^{-3t'}$ (thick red line) and the dimensionless covariance determinant for a number of sequences of \acp{CDM} (thinner lines), (b) Fitted law (dashed-pointed red) of a single \ac{CDM} sequence (dashed-pointed black).}
        \label{fig:fit_law}
    \end{figure}

    The approach described in previous sections assumes that every \ac{CDM} has the same relative importance and no additional source of information is available to qualify each individual \ac{CDM}. However, as the $t2TCA$ decreases, so does the effect of the uncertainty on the true shape of the distribution on the impact plane and the effect of model uncertainty in the propagation. \cref{fig:fit_law_a} shows the normalised determinant of multiple sequences of covariance matrices taken from the database of the \ac{ESA}'s Collision Avoidance Kelvins Challenge,  \cite{ESAChallenge2019,Uriot2022}. The database contains 13,152 sequences of \acp{CDM} of some of the \ac{LEO} satellites monitored by the ESA \ac{SDO}. 
    The figure shows that one can fit the simple exponential law $y'=e^{-3t'}$ to the magnitude of the determinant (red line in the figure). However, one cannot simply trust later \acp{CDM} due to large uncertainty in each individual sequence.
    Thus, we propose the following fit for each individual sequence:
    \begin{subequations}\label{eq:fit_law_norm_rule}
        \begin{equation}\label{eq:fit_law}
            y' = C e^{At'}+B\hspace{15pt} A,B,C\geq0,
        \end{equation}
        \begin{equation}\label{eq:dimensionless_cov}
            y' = \frac{\| \mathbf{\Sigma} \|}{\max_{CDMs}( \| \mathbf{\Sigma} \|)}
        \end{equation}
        \begin{equation}\label{eq:dimensionless_time}
            t' = \frac{(1-\max_{CDMs}(t2TCA))}{(\min_{CDMs}(t2TCA) - max_{CDMs}{t2TCA})}
        \end{equation}
    \end{subequations}
    
\begin{figure}[htb!]
        \centering
        \includegraphics[width=0.6\textwidth]{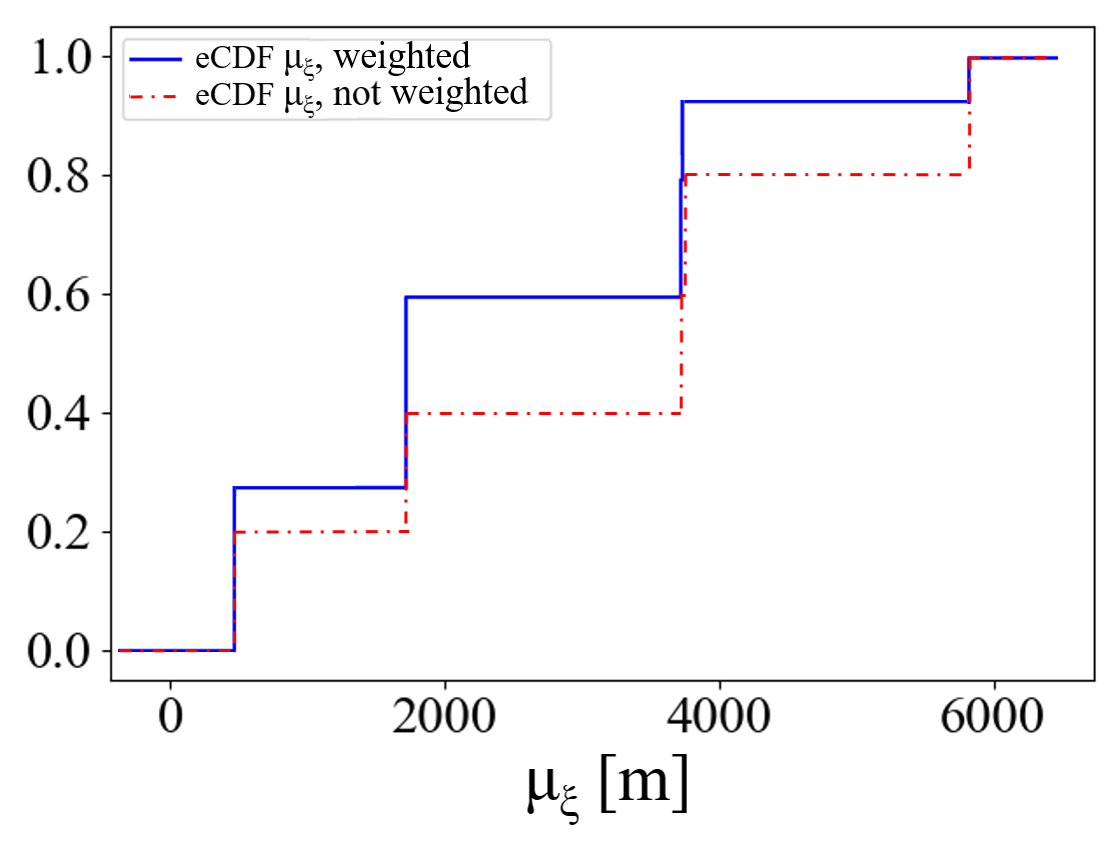}
        \captionsetup{width=0.8\linewidth,format=plain, font=small, labelfont=bf,justification=justified}
        \caption{eCDF for $\mu_\xi$ weighing the samples (blue) and with samples equally weighted (dashed red).}
        \label{fig:weighted_eCDF}
    \end{figure}

    Once the parameters $A$,$B$ and $C$ are fitted to the samples from a given sequence, the following weight is associated with each \ac{CDM} in that sequence:
    \begin{equation}\label{eq:bpa_weight}
        w_{CDM_i} = \frac{1}{y'(t2TCA_{CDM_i})}
    \end{equation}
    The weight is applied to each sample in the \ac{eCDF} used to compute the \ac{DKW} bounds: the probability mass associated with each sample is re-scaled by a factor $w_{CDM_i}$. See \cref{fig:weighted_eCDF} where the \ac{eCDF} of $\mu_\xi$ for an example with 5 observations is shown both with samples equally weighted (dashed red) or having applied the weighting law described above (blue).
    This approach results in a scaling of the probability mass associated with the \acp{CDM} but still allows the quantification of highly uncertain \acp{CDM} since there is no filtering process. The reason is that, with no information on trusted sources or individual \acp{CDM}, one cannot make any assumption on which \ac{CDM} is more credible.  

    \subsection{$\alpha$-cuts and DSt Structures}
    
\begin{figure}[htb!]
        \centering
        \includegraphics[width=\textwidth]{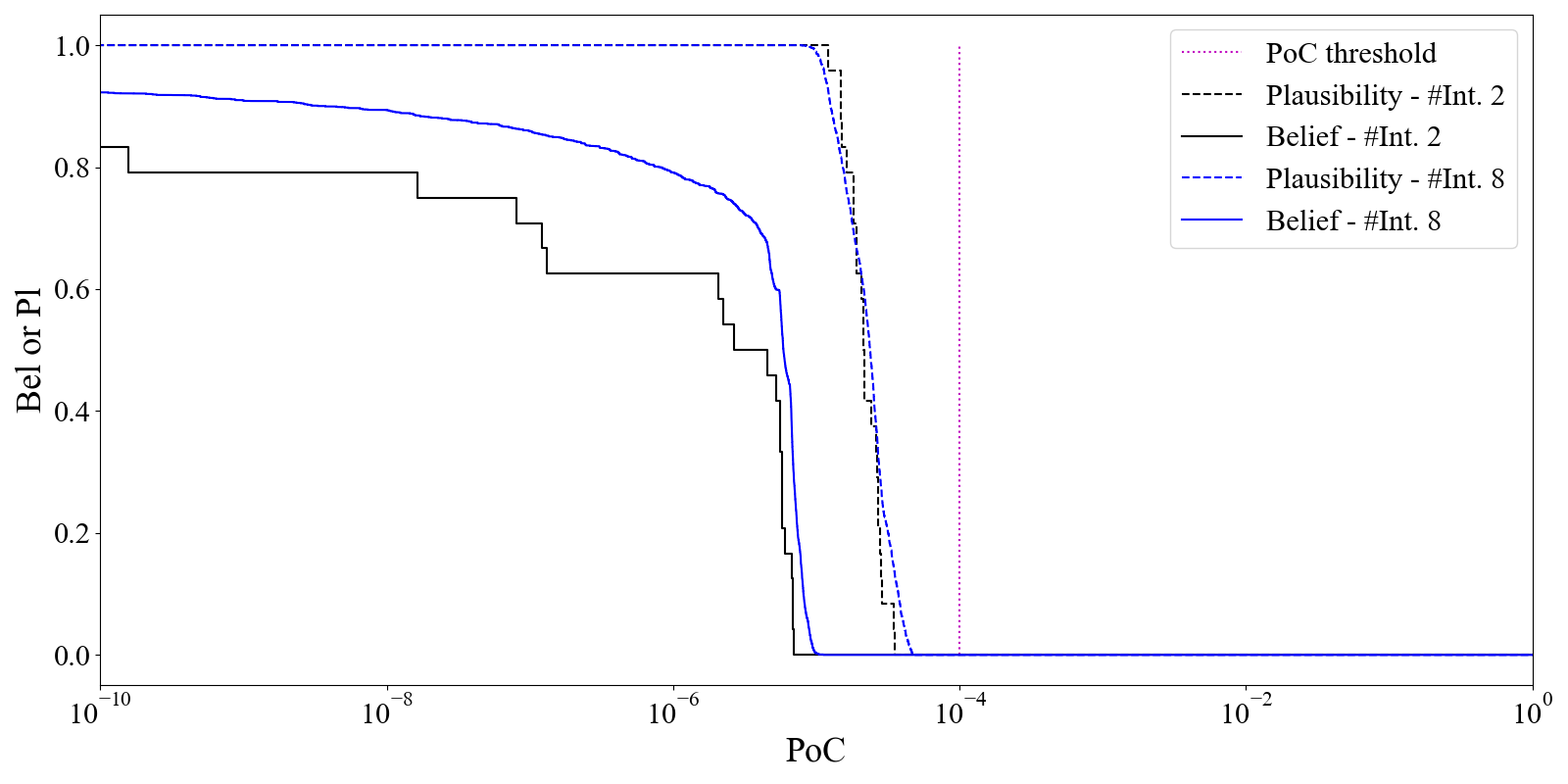}
        \captionsetup{width=0.8\linewidth,format=plain, font=small, labelfont=bf,justification=justified}
        \caption{Plausibility and Belief of $PoC\geq PoC_0$. Black: 1 $\alpha$-cut (two intervals) per variable, 32 FEs. Blue: 7 $\alpha$-cut (eight intervals) per variable, 32768 FEs. Solid lines: belief. Dashed lines: plausibility. Dotted purple vertical line: $PoC_0$.}
        \label{fig:PoC_CDM}
    \end{figure}

    Once a p-box is defined, the intervals for each component of $\mathbf{u}$ are derived from a series of equally spaced $\alpha$-cuts, light blue and grey dotted horizontal thin lines in \cref{fig:eCDF_d}. Each $\alpha$-cut creates interval, \cite{He2015,Chojnacki2015}: 
    \begin{equation}
        [x_\alpha,x^\alpha] = \left\{x\,\vert\,\mathcal{F}(x)\geq\alpha\right\}\text{.}
    \end{equation}
    The intersection with the upper bounds in the p-box defines the lower limit of the interval, and the intersections with the lower bound define the upper limit of the interval. The number of intervals is equal to the number of cuts plus one, and the \ac{bpa} associated with each interval, assuming the cuts are evenly spaced, is equal to the inverse of the number of cuts. The intervals and their \ac{bpa} will define an envelope around the p-box (blue and black dashed lines in \cref{fig:eCDF_d}). The greater the number of $\alpha$-cuts, the closer the envelope will be to the p-box, but the more computationally expensive is the computation of \ac{Bel} and \ac{Pl}. From the intervals associated with each component of $\mathbf{u}$ one can compute the FE $\gamma_i$ and their associated $bpa(\gamma_i)$ by performing the Cartesian product of all the intervals and associated \acp{bpa}. Once the \ac{FE} and \acp{bpa} are computed, the \ac{Pl}, \ac{Bel} of $PoC\geq PoC_0$ are computed with \cref{eq:BelPlOm} (see \cref{fig:PoC_CDM}) and the conjunction event is classified according to \cref{table:Crit}.
    
    Even in this case, we implicitly maintained the assumption that variables are independent, although it is not true that the components of the miss distance and of the covariance are all independent.  Approaches to address dependencies already exist in the literature, see \cite{Ferson2004}, and will be considered in future works. The independence assumption has two implications: i) the uncertainty space $U$ is an outer approximation of the space of all distributions of $\mathbf{u}$ and ii) some focal elements might not contain any sample of $\mathbf{u}$. The combination of the two generally leads to over-conservative results.
    Thus, in order to partially recover the interdependence between uncertain quantities, yet coherent with \ac{DSt}, a $bpa=0$ is assigned to all empty \acp{FE} and their \ac{bpa}, coming from the Cartesian product, is evenly distributed to the rest of \acp{FE} so that $\sum_i bpa(\gamma_i) = 1$.

%% file: 4_NumericalCases.tex
\section{Numerical experiments}
\label{sec:num_cases}

    In this section, some numerical tests are presented. The aim is to show the applicability of the methodology presented in previous sections and compare its outcome to the decisions made in past real cases by actual satellite operators: \ac{ESOC} and \ac{CNES}.

    \subsection{Parameter Tuning}
    \label{sec:threshold_cal}
    
    The methodology proposed in this paper requires the prior definition of the values of two thresholds: $Pl_0$ and $A_0$. These two thresholds should be tuned by analysing a large dataset of conjunction events with known outcomes. However, in every database of \acp{CDM} available to the authors, the number of provable \textit{Class 1} and \textit{2} conjunctions is very small or zero.

    % Spltting A and Pl threshold tuning
    Since $A_0$ does not affect \textit{Class 4} and \textit{5}, which depend only on $Pl_0$, but influences the number of \acp{TP} (actual collisions) and \acp{FP} (no-collisions believed to be collisions), one can define $Pl_0$ first and then use $A_0$ to quantify the degree of uncertainty in the class associated to an event. 
    % Expected outcome
    According to the classification in \cref{table:Crit}, the expected outcome is that low values of $Pl_0$ would increase the number of events classified as \textit{Class 1} or \textit{2}, reducing, at the same time, the number of \acp{FN} (collisions believed to be no-collisions) and increasing the amount of \acp{TP}. If this is combined with high values of $A_0$, the chances of detecting all high-risk events are high, but at the cost of increasing the number of \acp{FP}. If instead, $A_0$ is low, more events will be classified as uncertain (\textit{Class 0} and \textit{3}). On the contrary, a higher value of $Pl_0$ would reduce the false alerts, \acp{FP}, but at the risk of increasing the number of \acp{FN}.
    
    This paper used the \ac{DSt} structure to set a value for $Pl_0$. If there is at least one \ac{FE} supporting $PoC>PoC_0$, it means that there exists at least one piece of evidence suggesting that the \ac{PoC} can be correct. This piece of evidence may correspond to an extreme event with low probability. Following this idea, we propose the value $Pl_0 = \min_i(bpa(\gamma_i))$. This implies that even a \ac{PoC} that corresponds to a rare event in the generation of a \ac{CDM} is considered to be plausible.
    % Tuning the area threshold
    The value of $A_0$ is selected by balancing the number of \acp{TP} and \acp{FP}. The idea is to try to reduce the number of \acp{FP} by reclassifying them as uncertain cases and presenting the level of such uncertainty to the operator. A low value of $A_0$ implies that the operator accepts very little uncertainty in the sequence of \ac{CDM}, which reduces the number of \acp{FP} but potentially classifies some \acp{TP} as uncertain. On the other hand, a greater value of $A_0$ implies that the operator is very conservative and accepts to treat a number of \acp{FP} as \acp{TP}. 
    Thus, the decision to execute a \ac{CAM} is related to the confidence of the operator in the quality of the \acp{CDM}. For highly uncertain sequences of \acp{CDM}, a low $A_0$ is recommended, but if the quality of the \acp{CDM} is high, a higher $A_0$ should be used.

    % Area threshold tuning procedure: maximum area and A*
    In the following, rather than selecting the value of the area threshold $A_0$, we select the value of the normalised area $A^*_0 = \{0,0.05,0.1,0.15,...,0.95,1\}$, where $A^*_0$ is the fraction of the maximum possible area between the $Bel$ and $Pl$ curves, that is, when \ac{Bel} drops to zero at the minimum value of \ac{PoC}, $\underline{PoC}$, and \ac{Pl} remains equal to one until $PoC=1$. In this tuning exercise the area is computed by taking the lower limit $\underline{PoC}=10^{-30}$ for the \ac{PoC} as this is the lowest value computed from all the sequences of \acp{CDM} in our database. For all the first four tests in this paper, we will use a value of $A_0^*=0.1$ that allows one to clearly differentiate Event 1 from Events 3 and 4 in the following section. In the last test, we will present the sensitivity of the number of recommended \acp{CAM} to the value of $A_0^*$.

    \subsection{Comparison Against SDO and CNES}
    \label{sec:num_case_indiv}

    The results in this section will show a comparison between the \ac{CARA} performed with the proposed evidence-based method and the decisions made by real operators in a selected number of real cases. The two operators considered in this study are the \ac{ESA}'s \ac{SDO} and \ac{CNES}. Each of them has a different approach to conjunction analysis.  Four real conjunction events are analysed and 
    the different operational approaches are compared. 
    
\begin{figure}[htb!]
        \centering
        \includegraphics[width=\textwidth]{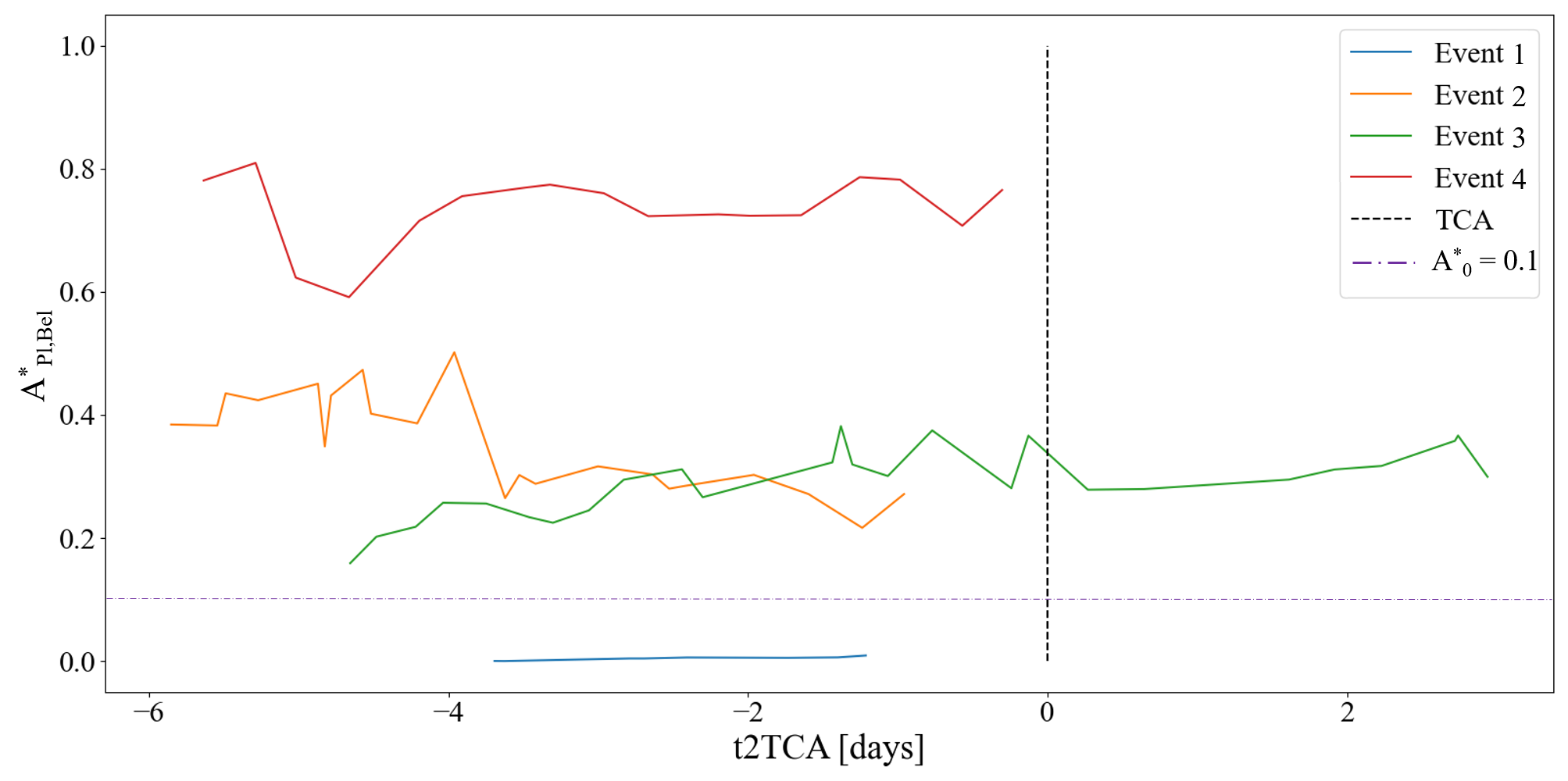}
        \captionsetup{width=0.8\linewidth,format=plain, font=small, labelfont=bf,justification=justified}
        \caption{Evolution of the normalised $A_{Pl,Bel}$ over time, for Events 1 to 4.}
        \label{fig:Area_threshold}
    \end{figure}

    For all examples the values of the thresholds are reported in in \cref{table:threshold}.
    \begin{table}[hbt!]
        \begin{center}
            %\small
            \captionsetup{width=0.8\linewidth,format=plain, font=small, labelfont=bf, justification=justified}
            \caption{Threshold values.}
            \begin{tabular}{ l l l}
                \toprule
                    \textbf{Threshold} & \textbf{Units} & \textbf{Value}\\
                \midrule
                    $T_1$ & days & 3 \\
                    $T_2$ & days & 5 \\
                    $PoC_0$ & - & $10^{-4}$ \\
                    $Pl_0$ & - & $1/\#FE$ \\
                    $A^*_0$ & - & $0.1$ \\
                    $\underline{PoC}$ & - & $10^{-30}$ \\
                \bottomrule
            \end{tabular}
            \label{table:threshold}
        \end{center}
    \end{table}
    The evolution of the normalised area gap between the $Pl$ and $Bel$ curves, or $A_{Pl,Brel}$, over time, for all four cases can be found in Figure \ref{fig:Area_threshold}, where $A^*_{Pl,Bel} = A_{Pl,Bel}/max(A_{Pl,Bel})$ is the normalised area between curves, $A_{Pl,Bel}$, defined in \cref{eq:BelPlOm_area}. The Figure confirms that an $A^*_0=0.1$ is appropriate to differentiate between cases like Event 1 from cases like Event 3 and 4. All four cases are presented in more detail in the following subsections.

    \subsubsection{Space Debris Office Conjunction Risk Assessment}
    The approach followed by the \ac{ESA} \ac{SDO} is probability-based, relying mainly on the value of the \ac{PoC} computed with the information from the \acp{CDM}, or the \ac{PoC} included in the \ac{CDM}. The following quote may summarise the conjunction risk assessment process adopted by the \ac{SDO}: \say{For a given close approach the last obtained CDM, including the computed risk, can be assumed to be the best knowledge we have about the potential collision and the state of the two objects in question. In most cases, the Space Debris Office will alarm control teams and start thinking about a potential avoidance manoeuvre 2 days prior to the close approach in order to avoid the risk of collision, to then make a final decision 1 day prior},  \cite{ESAChallenge2019}. Nevertheless, each mission monitored by the \ac{SDO} has specific operational constraints (i.e. the time needed to prepare and execute the manoeuvre) and will have its own risk and time thresholds, $PoC_0$ and $T_1$. The time threshold $T_1$ is generally 2 or 3 days away from TCA. At that point the mission team is informed about the possible collision, and a final decision is usually made (when possible) 1 day from TCA,  \cite{ESAChallenge2019}. 
    The risk threshold $PoC_0$ is determined statistically based on the overall collision risk and the annual frequency of close approaches, trading off the ignored risk and the avoided risk by selecting the risk threshold at the cost of an expected number of annual manoeuvres, see \cite{Merz2017}. Generally, for missions in the \ac{LEO} regime, a threshold of $PoC_0=10^{-4}$ leads to a risk reduction of around 90\% at the expense of 1 to 3 manoeuvres per year, with current levels of traffic. However, a lower threshold, around $10^{-5}$, may be considered to ensure sufficient time to prepare a collision avoidance manoeuvre in the case of escalated events,  \cite{Merz2017}. 
    
    Following this approach, the \ac{SDO} escalates an event when the \ac{PoC} of the last \ac{CDM} is bigger than the threshold. Escalating an event means that further and more detailed analyses are required. If the risk is still above the threshold at the decision time, a \ac{CAM} is designed in cooperation with the mission team, whose final decision will be made based on the value of \ac{PoC} included in the last \ac{CDM} received before the go/no-go decision time.
    More detailed information on the CARA process of the \ac{SDO} can be found in \cite{Merz2017}.
    For the first three events in this subsection, only \acp{CDM}s from the MiniCat database were considered.

    \subsubsubsection{Event \#1}
    
\begin{figure}[htb!]
        \centering
        \begin{subfigure}[b]{0.49\textwidth}
            \centering
            \includegraphics[width=\textwidth]{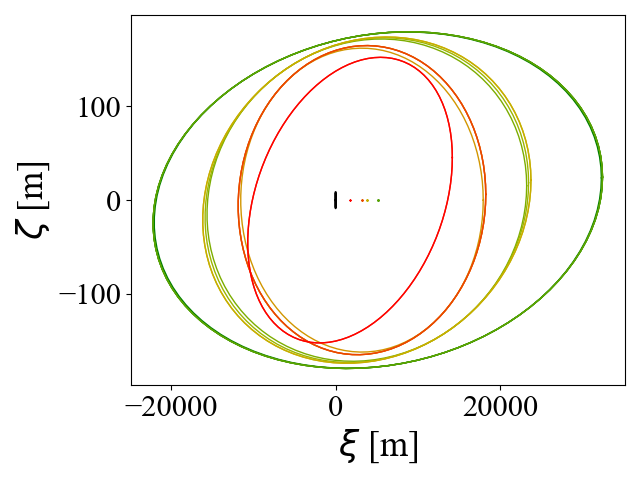}
            \caption{}
            \label{fig:CDM_NC2_Sc_1_a}
        \end{subfigure}
        \begin{subfigure}[b]{0.49\textwidth}
            \centering
            \includegraphics[width=\textwidth]{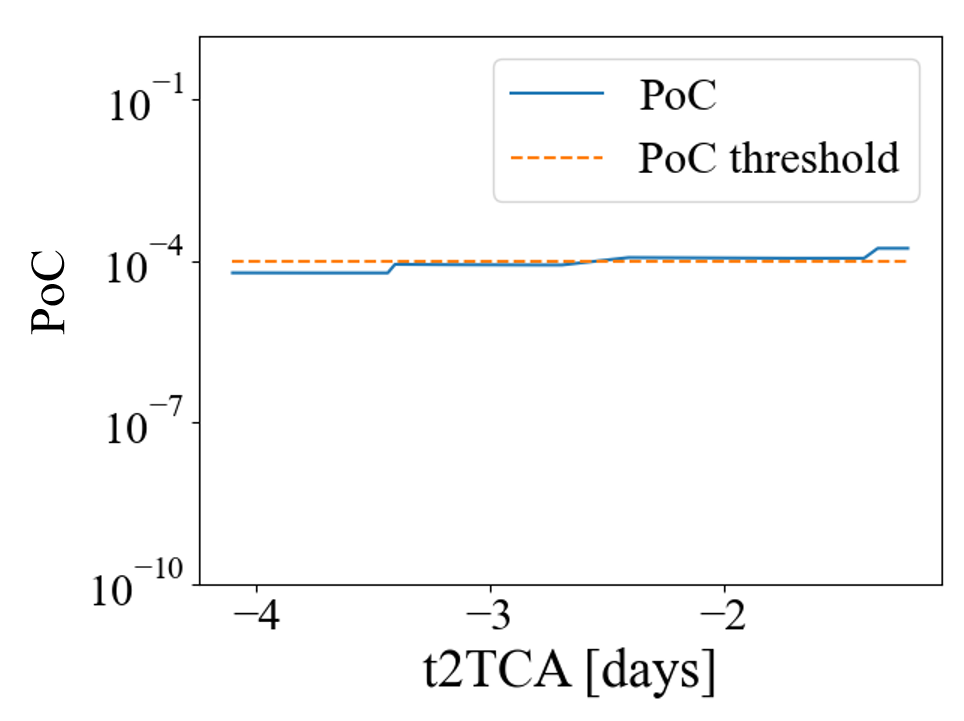}
            \caption{}
            \label{fig:CDM_NC2_Sc_1_b}
        \end{subfigure}
        \captionsetup{width=0.8\linewidth,format=plain, font=small, labelfont=bf,justification=justified}
        \caption{CDM information for example in Event \#1: High-risk event. (a) Uncertain ellipses in the sequence of CDMs. Green ellipses correspond to earlier CDMs, and red ellipses to later CDMs. (b) Evolution of the PoC in the CDMs with the time to the TCA. Blue solid line: PoC; orange dashed line: PoC threshold.} 
        \label{fig:CDM_NC2_Sc_1}
    \end{figure}

    This event represents a high-risk scenario provided by the \ac{ESA} \ac{SDO}. The uncertain geometry in the impact plane, with the whole sequence of \acp{CDM} and the \ac{PoC} evolution are displayed in \cref{fig:CDM_NC2_Sc_1}. Events with \ac{PoC} above the threshold for times to \ac{TCA} greater than $T_1$ make the event escalate, that is, they are further analysed and possible alerts to the mission's team can be triggered, while high-risk \acp{CDM} received in the last 72 hours trigger a \ac{CAM} procedure.
    
    From \cref{fig:CDM_NC2_Sc_1_b}, one can see that the \ac{PoC} remains high along the whole sequence. Even if at the beginning it was below the threshold, its proximity to $PoC_0$ along with the upward trend made the operator escalate the event. The \ac{PoC} threshold was violated within the last few days before TCA, which led to a \ac{CAM} execution to reduce the risk of the event.
    
\begin{figure}[htb!]
        \centering
        \begin{subfigure}[b]{0.49\textwidth}
            \centering
            \includegraphics[width=\textwidth]{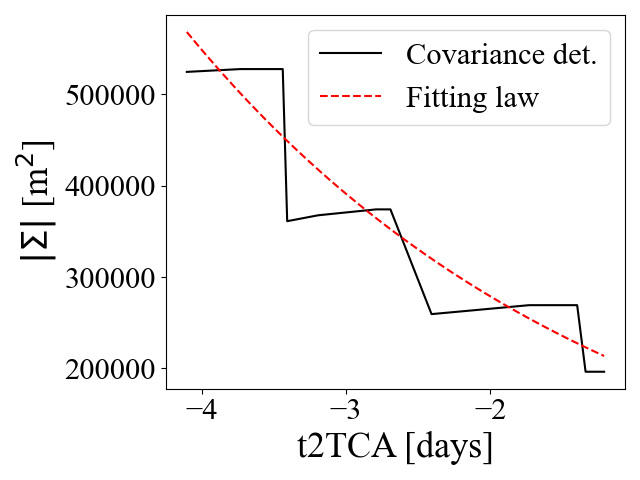}
            \caption{}
            \label{fig:Fitlaw_NC2_Sc_1_a}
        \end{subfigure}
        \begin{subfigure}[b]{0.49\textwidth}
            \centering
            \includegraphics[width=\textwidth]{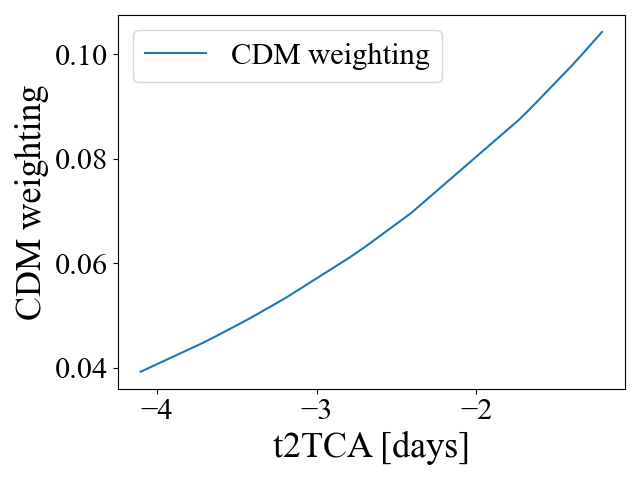}
            \caption{}
            \label{fig:Fitlaw_NC2_Sc_1_b}
        \end{subfigure}
        \captionsetup{width=0.8\linewidth,format=plain, font=small, labelfont=bf,justification=justified}
        \caption{Fitting law to weight the CDMs after having received the whole sequence in Event \#1: High-risk event. (a) Solid black line: value of the determinant from the CDMs, dashed red line: fitting law of the covariance matrix determinant. (b) Weight of the CDMs as a function of the time to the TCA.} 
        \label{fig:Fitlaw_NC2_Sc_1}
    \end{figure}

    We applied our evidence-based methodology to this case by following the approach presented in \cref{sec:epist_CDM}. The \ac{DKW} bands were computed assuming a confidence interval $\delta=0.5$. The \acp{CDM} were weighted according to the exponential law in \cref{eq:fit_law_norm_rule}. \cref{fig:Fitlaw_NC2_Sc_1} shows the fitting law after having received all the \acp{CDM} (red) along with the value of the combined covariance matrix determinant, for the whole sequence (black). For the fitting law in \cref{fig:Fitlaw_NC2_Sc_1_b}, the value of the dimensionless parameters in \cref{eq:fit_law} after having received the whole sequence are: $A = 1.0752, B=0.9811, C=0.001716$. Note that the value of the parameters varies with the number of \acp{CDM} received to better fit the covariance determinant evolution up to that time.
    
\begin{figure}[htb!]
        \centering
        \includegraphics[width=\textwidth]{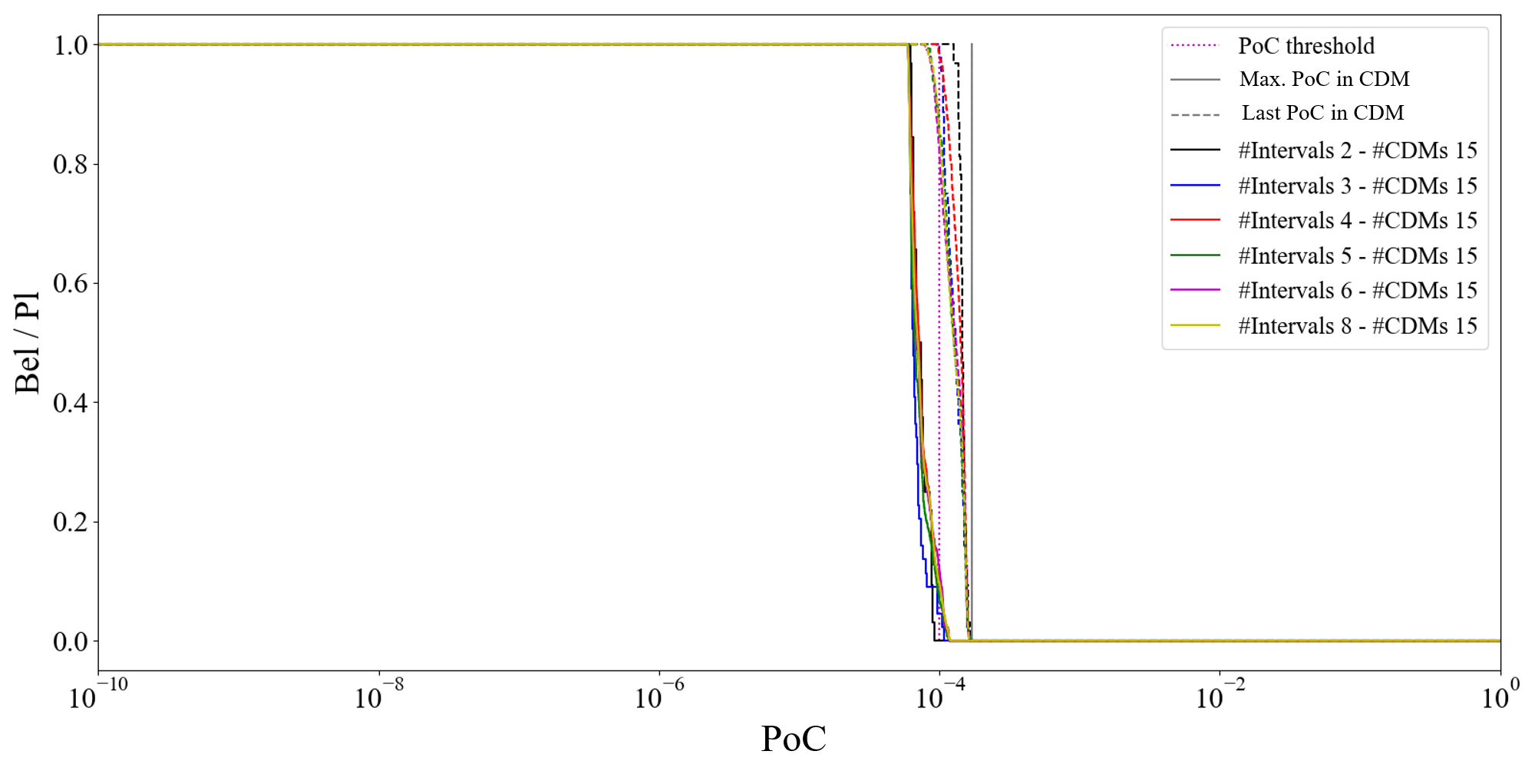}
        \captionsetup{width=0.8\linewidth,format=plain, font=small, labelfont=bf,justification=justified}
        \caption{\textit{Pl} and \textit{Bel} of the PoC after having received the whole sequence of CDMs Event \#1: High-risk event. Solid vertical grey line: maximum PoC in the sequence, dashed vertical grey line: PoC of last CDM, pointed purple line: PoC threshold. For the rest of the colours: Belief in solid lines and Plausibility in dashed lines. Black: 1 $\alpha$-cut per variable (2 intervals per variable, 32 FEs), blue: 2 $\alpha$-cuts, red: 3 $\alpha$-cuts, green: 4 $\alpha$-cuts, purple: 5 $\alpha$-cuts, yellow: 7$\alpha$-cuts.} 
        \label{fig:PlBel_NC2_Sc_1}
    \end{figure}

    We repeated the same analysis with different numbers of $\alpha$-cuts per uncertain variable: $\#\alpha\text{-cuts} = \left\{1,2,3,4,5,7\right\}$. These cuts led to a number of intervals per variable equal to $\#\text{intervals} = \left\{2,3,4,5,6,8\right\}$, which translated into a number of \acp{FE} $\#\text{FE} 
    = \left\{32,243,1024,3125,7776,16807\right\}$, respectively. The \ac{Pl} and \ac{Bel} curves for the \ac{PoC}, for each number of cuts, is presented in \cref{fig:PlBel_NC2_Sc_1}, after having received the whole sequence of \acp{CDM}.
    
    \cref{fig:PlBel_NC2_Sc_1} shows that, although the increasing number of $\alpha$-cuts provides a more refined set of curves, their shape and values varies only slightly. In this case, the \ac{Bel} and \ac{Pl} curves overlap for most values of \ac{PoC} except for a small interval around the $PoC_0$, as it could be expected both, from the uncertainty geometry in \cref{fig:CDM_NC2_Sc_1_a} and the values of the \ac{PoC} in \cref{fig:CDM_NC2_Sc_1_b}. Since the information in the \ac{CDM} is coherent across the whole sequence, the gap between the \ac{Pl} and \ac{Bel} curves is small.  
    
\begin{figure}[htb!]
        \centering
        \includegraphics[width=\textwidth]{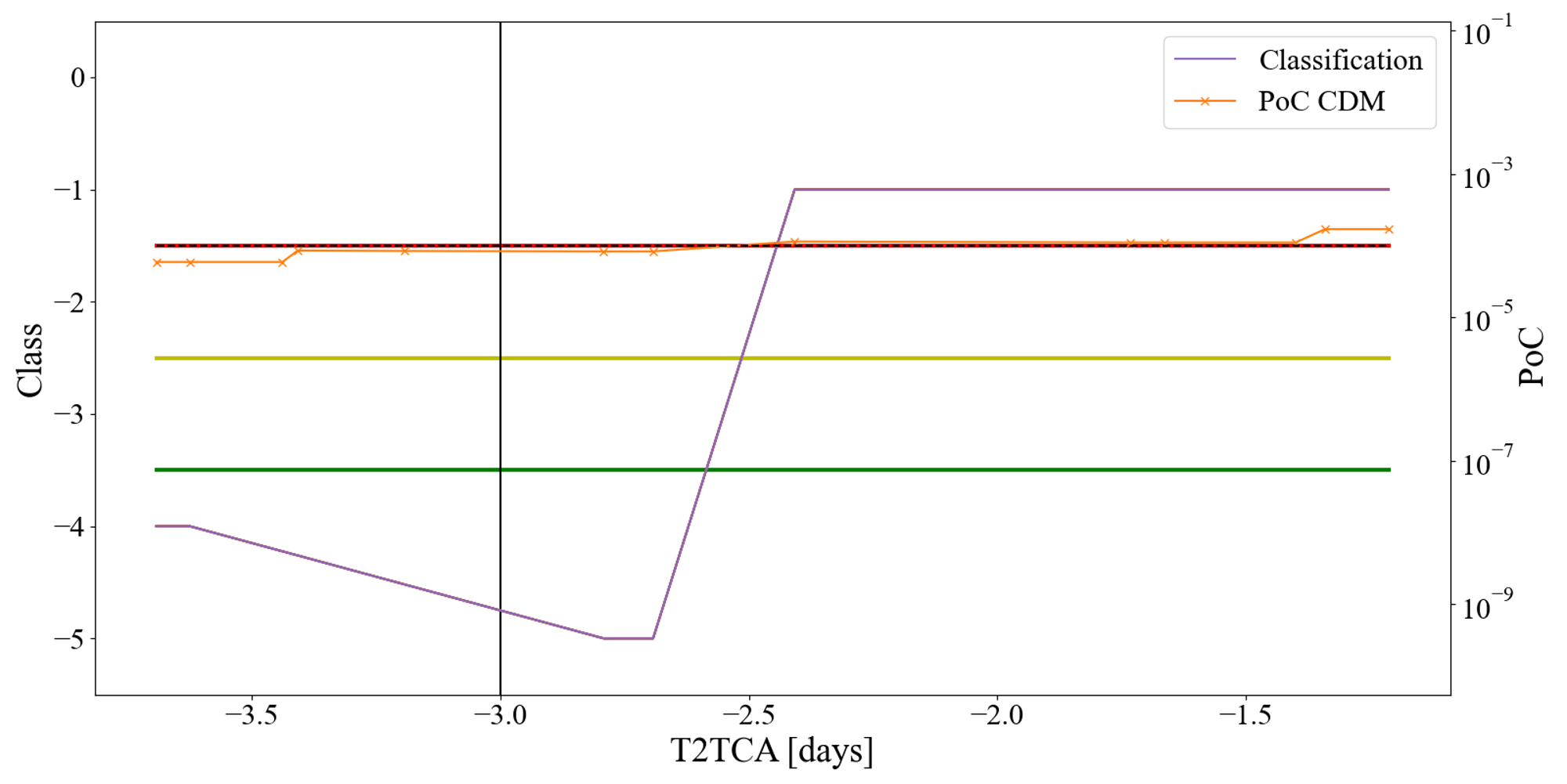}
        \captionsetup{width=0.8\linewidth,format=plain, font=small, labelfont=bf,justification=justified}
        \caption{Collision risk assessment for Event \#1: High-risk event. Solid narrow lines: evidence-based classification with different number of $\alpha$-cuts: $\#\alpha\text{-cuts}=\left\{1,2,3,4,5,7\right\}$ (note that they overlap each other, so only $\#\alpha\text{-cuts}=7$ is visible in solind purple). Crossed-solid line: PoC in the CDMs used by SDO for assessment. Horizontal thick lines: evidence approach safety bands: green, low risk-uncertain boundary; yellow, uncertain-high risk boundary; red, mid term high risk-long term high risk boundary. Dashed black line: Risk threshold (overlapping evidence-based high-risk boundary). Vertical black line: decision time threshold $T_1$. }
        \label{fig:Clas_NC2_Sc_1_seq}
    \end{figure}

    \cref{fig:Clas_NC2_Sc_1_seq} shows the classification, purple solid line, as a function of the time to the \ac{TCA} from the last received \ac{CDM}. The figure shows also the \ac{PoC} directly computed from the \ac{CDM}.
    
    Initially, the event is classified as \textit{Class 4} and rapidly falls to \textit{Class 5}, since there is little evidence supporting a higher \ac{PoC}. However, at 2.5 days from TCA, the \ac{PoC} consistently grows above the threshold. Given the little uncertainty in the sequence of \acp{CDM} the event is reclassified as \textit{Class 1} and a \ac{CAM} is recommended.
    
    This is the same decision finally taken by the \ac{SDO}. As seen in \cref{fig:PlBel_NC2_Sc_1}, the support for a high value of \ac{PoC} is high and the gap between the curves (level of uncertainty) is very small. Thus, the recommended action in the last days prior to the encounter would be to implement a manoeuvre to reduce the risk of a collision.
    
    \subsubsubsection{Event \#2}
    
\begin{figure}[htb!]
        \centering
        \begin{subfigure}[b]{0.49\textwidth}
            \centering
            \includegraphics[width=\textwidth]{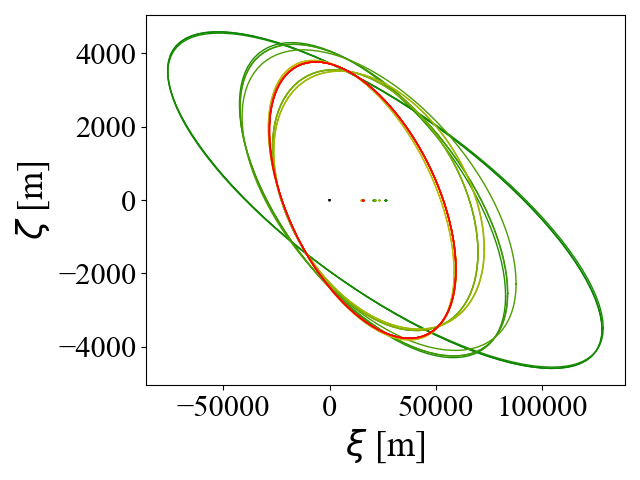}
            \caption{}
            \label{fig:CDM_NC2_Sc_2_a}
        \end{subfigure}
        \begin{subfigure}[b]{0.49\textwidth}
            \centering
            \includegraphics[width=\textwidth]{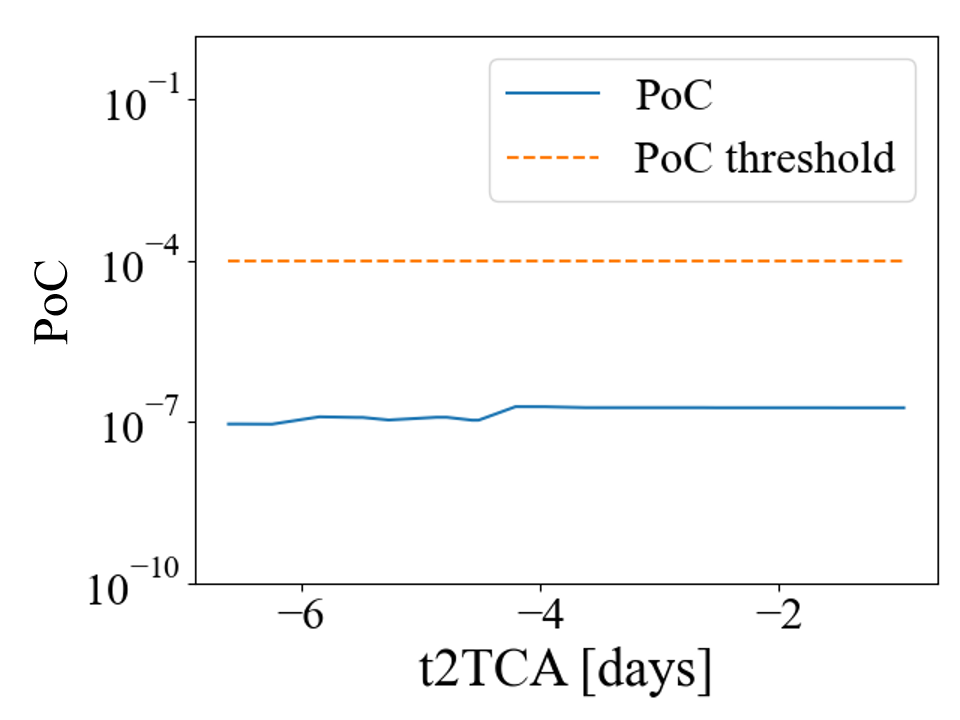}
            \caption{}
            \label{fig:CDM_NC2_Sc_2_b}
        \end{subfigure}
        \captionsetup{width=0.8\linewidth,format=plain, font=small, labelfont=bf,justification=justified}
        \caption{CDM information for example in Event \#2: Low-risk event. (a) Uncertain ellipses in the sequence of CDMs. Green ellipses correspond to earlier CDMs, and red ellipses to later CDMs. (b) Evolution of the PoC in the CDMs with the time to the TCA. Blue solid line: PoC; orange dashed line: PoC threshold.} 
        \label{fig:CDM_NC2_Sc_2}
    \end{figure}

    A similar analysis was done for the Low-risk conjunction event illustrated in \cref{fig:CDM_NC2_Sc_2}, also provided by the \ac{ESA} \ac{SDO}. Opposite to the previous event, in this case, the \ac{PoC} remains well below the threshold, so no alert is required to be triggered and no \ac{CAM} is required to be designed or executed.
    
\begin{figure}[htb!]
        \centering
        \begin{subfigure}[b]{0.49\textwidth}
            \centering
            \includegraphics[width=\textwidth]{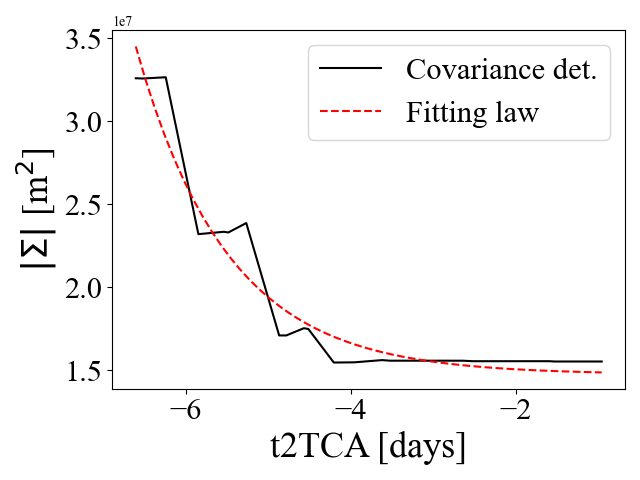}
            \caption{}
            \label{fig:Fitlaw_NC2_Sc_2_a}
        \end{subfigure}
        \begin{subfigure}[b]{0.49\textwidth}
            \centering
            \includegraphics[width=\textwidth]{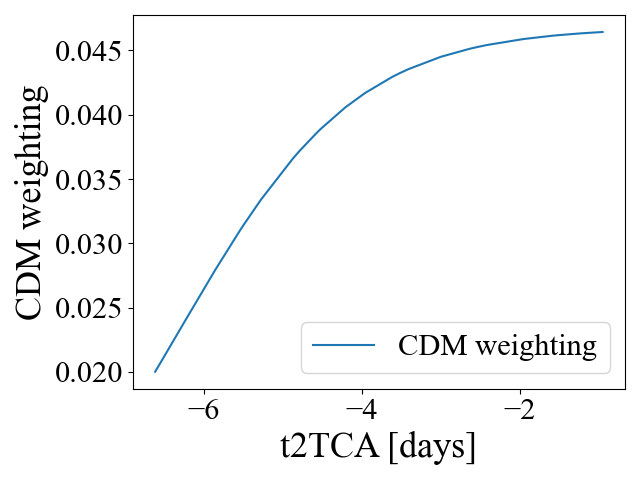}
            \caption{}
            \label{fig:Fitlaw_NC2_Sc_2_b}
        \end{subfigure}
        \captionsetup{width=0.8\linewidth,format=plain, font=small, labelfont=bf,justification=justified}
        \caption{Fitting law to weight the CDMs after having received the whole sequence in Event \#2: Low-risk event. (a) Solid black line: value of the determinant from the CDMs, dashed red line: fitting law of the covariance matrix determinant. (b) Weight of the CDMs as a function of the time to the TCA.} 
        \label{fig:Fitlaw_NC2_Sc_2}
    \end{figure}

    The evidence-based analysis was performed using the same parameters as before: $\delta=0.5$ for the \ac{DKW} bands, with a different number of $\alpha$-cuts: $\#\alpha\text{-cuts}=\left\{1,2,3,4,5,7\right\}$ per variable. The final set of \acp{CDM} was weighted with the exponential fitting law \cref{eq:fit_law} using the following parameters: $A = 0.6049, B=5.0896, C=0.4518$. The fitting law (red) and the combined covariance matrix determinant in the \acp{CDM} (black) appear in \cref{fig:Fitlaw_NC2_Sc_2}. Note the convergence in the second half of the sequence.
    
\begin{figure}[htb!]
        \centering
        \includegraphics[width=\textwidth]{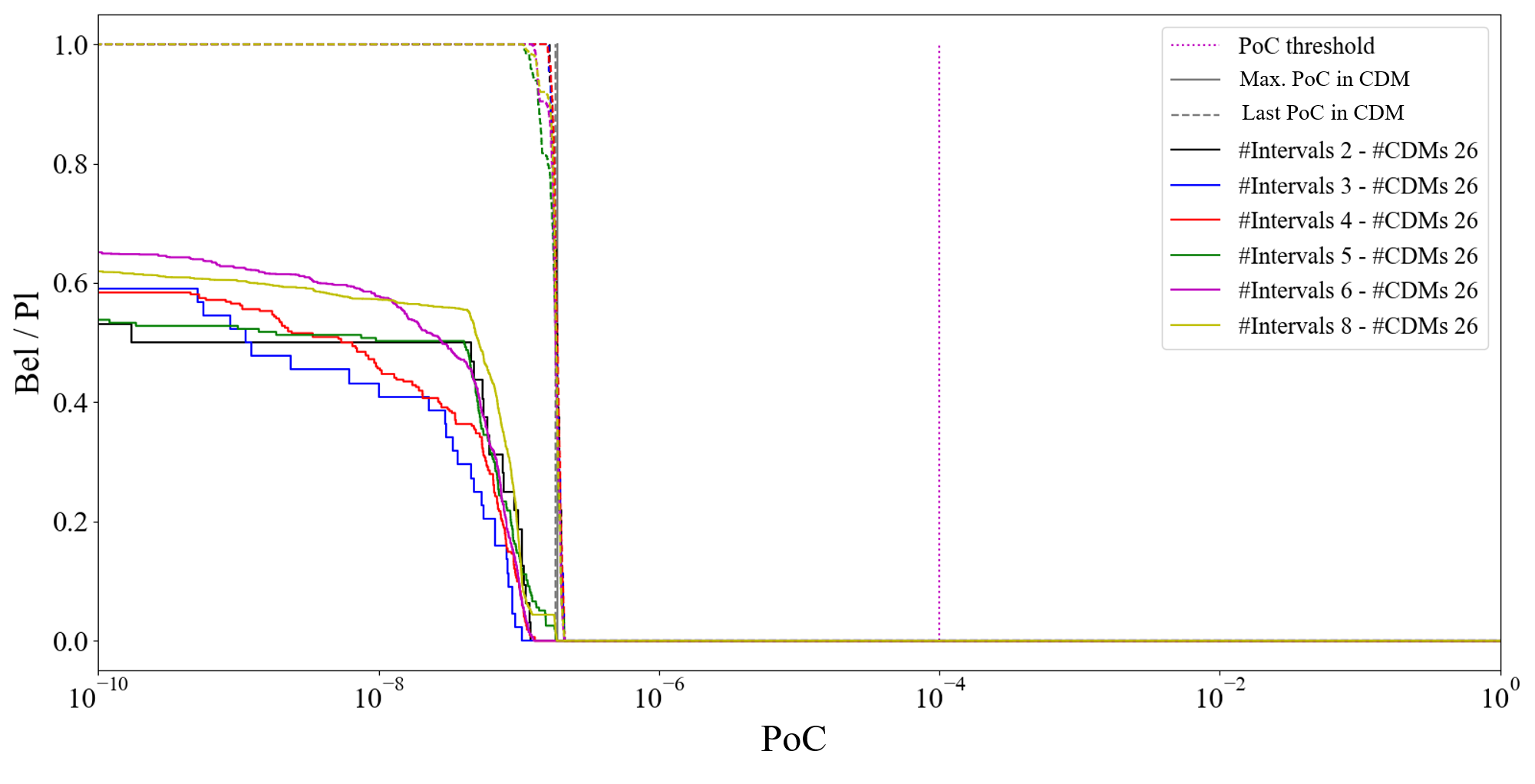}
        \captionsetup{width=0.8\linewidth,format=plain, font=small, labelfont=bf,justification=justified}
        \caption{\textit{Pl} and \textit{Bel} of the PoC after having received the whole sequence of CDMs Event \#2: Low-risk event. Solid vertical grey line: maximum PoC in the sequence, dashed vertical grey line: PoC of last CDM, pointed purple line: PoC threshold. For the rest of the colours: Belief in solid lines and Plausibility in dashed lines. Black: 1 $\alpha$-cut per variable (2 intervals per variable, 32 FEs), blue: 2 $\alpha$-cuts, red: 3 $\alpha$-cuts, green: 4 $\alpha$-cuts, purple: 5 $\alpha$-cuts, yellow: 7$\alpha$-cuts.} 
        \label{fig:PlBel_NC2_Sc_2}
    \end{figure}

    In \cref{fig:PlBel_NC2_Sc_2}, the corresponding \ac{Pl} and \ac{Bel} curves on the value of \ac{PoC} after having received all the \acp{CDM} of the event are shown. Again, increasing the number of $\alpha$-cuts makes the curves smoother and shows a converging trend, but does not change the overall confidence in the value \ac{PoC}. The maximum value of \ac{PoC} with some supporting evidence is well below the threshold, indicating that the event can be deemed to be safe. However, the left-most part of the \ac{Bel} and \ac{Pl} curves shows a significant gap. This can be explained by the fact that the ellipses are not too different from each other (\cref{fig:CDM_NC2_Sc_2_a}) and they tend to converge to a single ellipse for the later \acp{CDM}, as shown in \cref{fig:CDM_NC2_Sc_2_b}. Thus, the initial information content in each \ac{CDM}tends to support lower values of \ac{PoC}, which explains the lower value of \ac{Bel} on the left of the graph. However, due to the concentration of information around the later \acp{CDM}, the big drop both in \ac{Pl} and \ac{Bel} occurs at \ac{PoC}$\sim 10^{-7}$.

\begin{figure}[htb!]
        \centering
        \includegraphics[width=\textwidth]{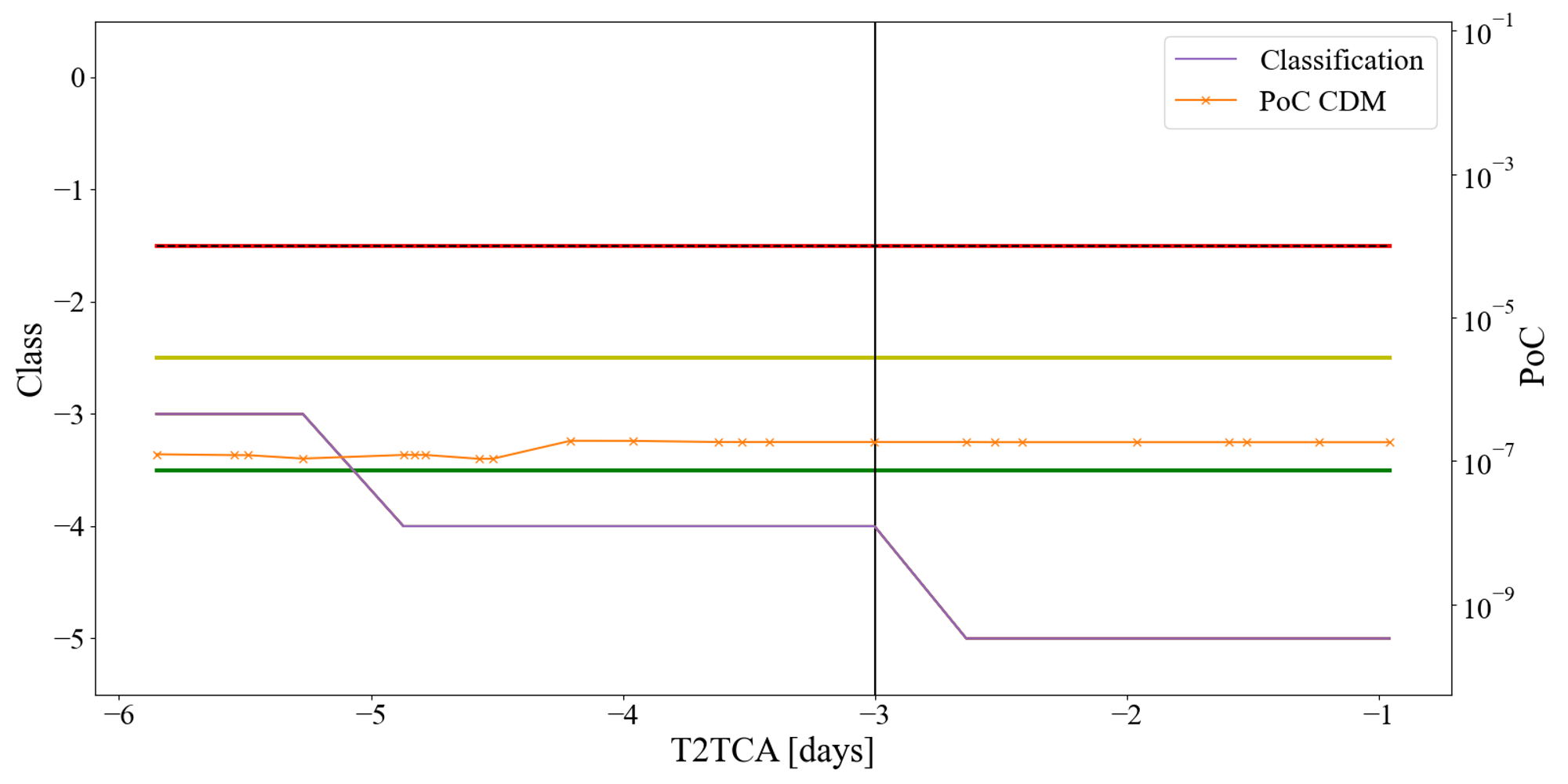}
        \captionsetup{width=0.8\linewidth,format=plain, font=small, labelfont=bf,justification=justified}
        \caption{Collision risk assessment for Event \#2: Low-risk event. Solid narrow lines: evidence-based classification with different number of $\alpha$-cuts: $\#\alpha\text{-cuts}=\left\{1,2,3,4,5,7\right\}$ (note that they overlap each other, so only $\#\alpha\text{-cuts}=7$ is visible in solid purple). Crossed-solid line: PoC in the CDMs used by SDO for assessment. Horizontal thick lines: evidence approach safety bands: green, low risk-uncertain boundary; yellow, uncertain-high risk boundary; red, mid term high risk-long term high risk boundary. Dashed black line: Risk threshold (overlapping evidence-based high-risk boundary). Vertical black line: decision time threshold.}
        \label{fig:Clas_NC2_Sc_2_seq}
    \end{figure}

    Finally, the conjunction assessment for the whole sequence is shown in \cref{fig:Clas_NC2_Sc_2_seq}. This event displays a greater uncertainty with respect to the previous scenario, but values of the \ac{PoC} greater than $10^{-7}$ have no supporting evidence and $Pl=Bel=0$. Thus, the event is initially classified as \textit{Class 4} ($t2TCA>T_1$) and then dropped to \textit{Class 5} ($t2TCA\leq T_1$) for the whole sequence,
    meaning that no further action should be taken by the operator. 
    This is the same decision made by the \ac{SDO}.
    
    \subsubsubsection{Event \#3}
    
\begin{figure}[htb!]
        \centering
        \begin{subfigure}[b]{0.49\textwidth}
            \centering
            \includegraphics[width=\textwidth]{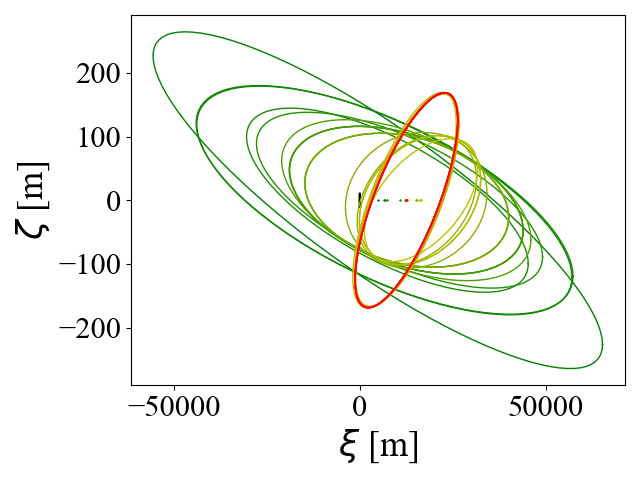}
            \caption{}
            \label{fig:CDM_NC2_Sc_3_a}
        \end{subfigure}
        \begin{subfigure}[b]{0.49\textwidth}
            \centering
            \includegraphics[width=\textwidth]{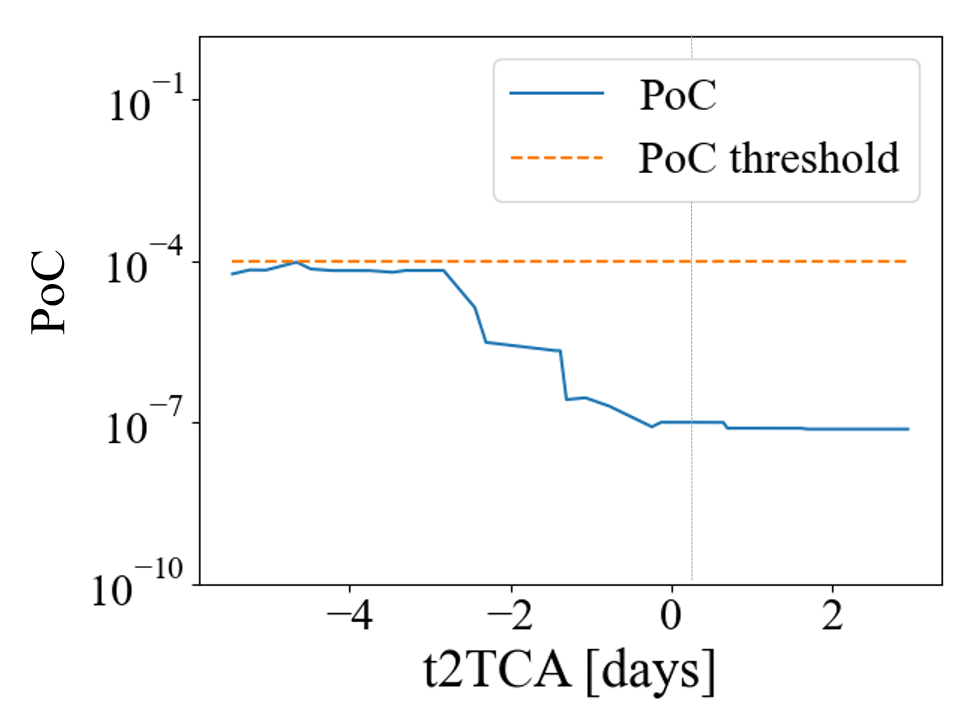}
            \caption{}
            \label{fig:CDM_NC2_Sc_3_b}
        \end{subfigure}
        \captionsetup{width=0.8\linewidth,format=plain, font=small, labelfont=bf,justification=justified}
        \caption{CDM information for example in Event \#3: Uncertain event. (a) Uncertain ellipses in the sequence of CDMs. Green ellipses correspond to earlier CDMs, and red ellipses to later CDMs. (b) Evolution of the PoC in the CDMs with the time to the TCA. Blue solid line: PoC; orange dashed line: PoC threshold, vertical dashed grey line: TCA.} 
        \label{fig:CDM_NC2_Sc_3}
    \end{figure}

    This last event is affected by a significant level of uncertainty.
    The encounter geometry and the evolution of the \ac{PoC} in the \acp{CDM} are shown in \cref{fig:CDM_NC2_Sc_3}. Despite the initial high risk, with values of \ac{PoC} close to the threshold, the final decision of the \ac{SDO} was not to take any further action. This decision was dictated by the later values of the \ac{PoC}, that were all consistently lower than the initial ones, and considerably below $PoC_0$.
    
\begin{figure}[htb!]
        \centering
        \begin{subfigure}[b]{0.49\textwidth}
            \centering
            \includegraphics[width=\textwidth]{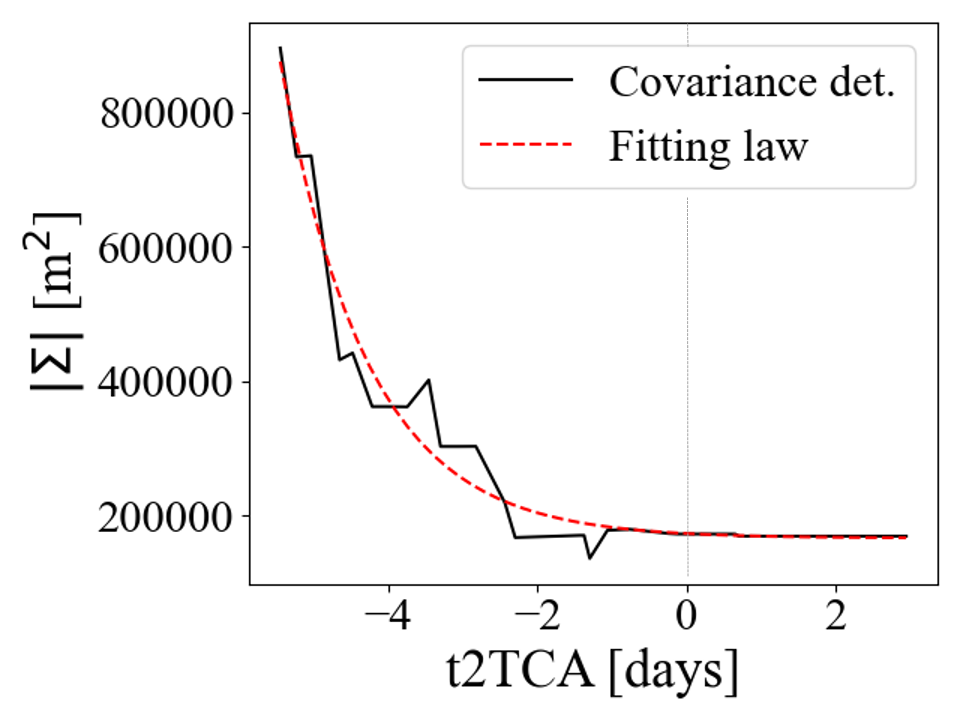}
            \caption{}
            \label{fig:Fitlaw_NC2_Sc_3_a}
        \end{subfigure}
        \begin{subfigure}[b]{0.49\textwidth}
            \centering
            \includegraphics[width=\textwidth]{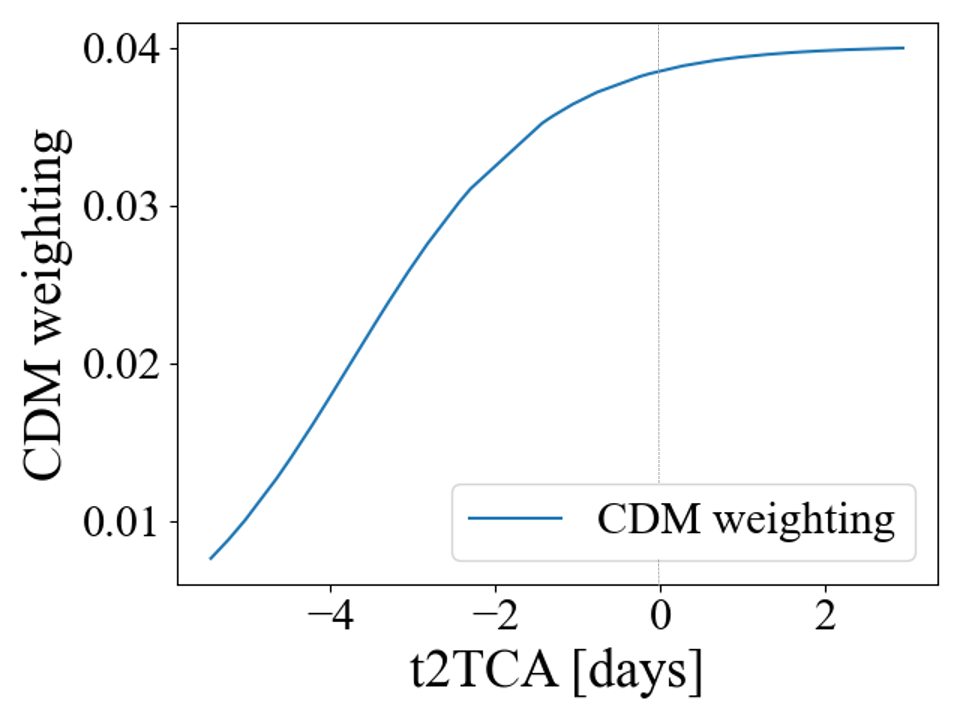}
            \caption{}
            \label{fig:Fitlaw_NC2_Sc_3_b}
        \end{subfigure}
        \captionsetup{width=0.8\linewidth,format=plain, font=small, labelfont=bf,justification=justified}
        \caption{Fitting law to weight the CDMs after having received the whole sequence in Event \#3: Uncertain event. (a) Solid black line: value of the determinant from the CDMs, dashed red line: fitting law of the covariance matrix determinant. Vertical dashed grey line: TCA. (b) Weight of the CDMs as a function of the time to the TCA. Vertical dashed grey line: TCA.} 
        \label{fig:Fitlaw_NC2_Sc_3}
    \end{figure}

    The evidence-based analysis was performed with the same parameters as before: $\delta=0.5$ for the \ac{DKW} bands. 
    The exponential fitting law \cref{eq:fit_law} to weight the \acp{CDM}, after having received the whole sequence, had the following parameters $A=0.7917,B=7.1471,C=0.1858$ and is shown in \cref{fig:Fitlaw_NC2_Sc_3} (red line) along with the covariance matrix determinant (black line). 
    
\begin{figure}[htb!]
        \centering
        \includegraphics[width=\textwidth]{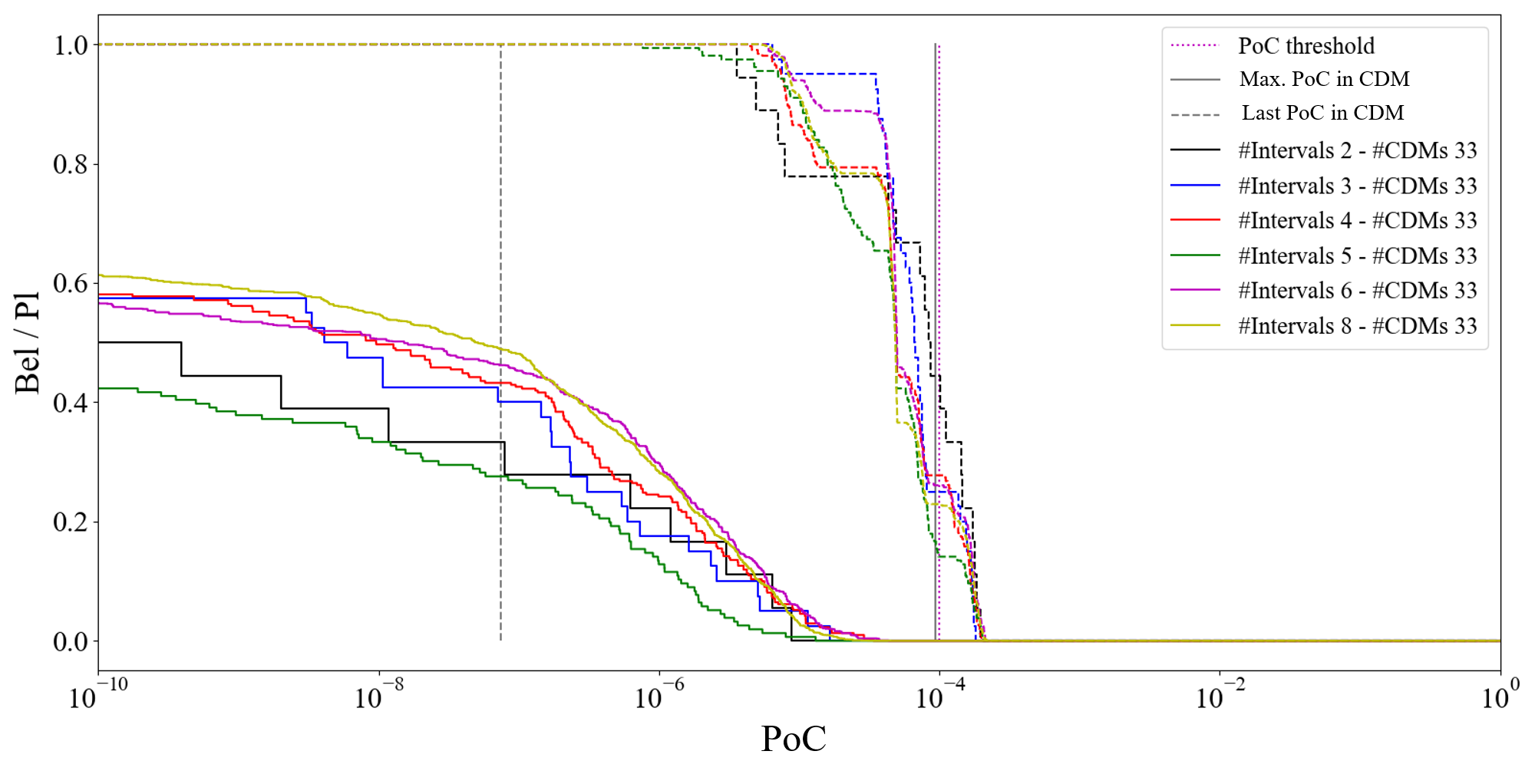}
        \captionsetup{width=0.8\linewidth,format=plain, font=small, labelfont=bf,justification=justified}
        \caption{\textit{Pl} and \textit{Bel} of the PoC after having received the whole sequence of CDMs Event \#3: Uncertain event. Solid vertical grey line: maximum PoC in the sequence, dashed vertical grey line: PoC of last CDM, pointed purple line: PoC threshold. For the rest of the colours: Belief in solid lines and Plausibility in dashed lines. Black: 1 $\alpha$-cut per variable (2 intervals per variable, 32 FEs), blue: 2 $\alpha$-cuts, red: 3 $\alpha$-cuts, green: 4 $\alpha$-cuts, purple: 5 $\alpha$-cuts, yellow: 7$\alpha$-cuts.} 
        \label{fig:PlBel_NC2_Sc_3}
    \end{figure}

    The \ac{Pl} and \ac{Bel} curves for the \ac{PoC} were computed for different $\alpha$-cuts: $\#\alpha\text{-cuts}=\left\{1,2,3,4,5,7\right\}$. The curves are shown in \cref{fig:PlBel_NC2_Sc_3}. In this case, there is a significant gap between \ac{Pl} and \ac{Bel} for all the values of \ac{PoC} for which $Pl>0$. 
    This uncertainty (or level of disagreement between \acp{CDM}) can be seen in \cref{fig:CDM_NC2_Sc_3_a}, which shows the variety of the uncertainty ellipses from the beginning of the sequence to the last \acp{CDM}. In this case the supporting evidence that a value of $PoC>PoC_0$ is plausible does not go to zero but the gap between the \ac{Pl} and \ac{Bel} curves suggests that a further analysis is required although the value of \ac{Pl} is low and \ac{Bel} is zero.  

\begin{figure}[htb!]
        \centering
        \includegraphics[width=\textwidth]{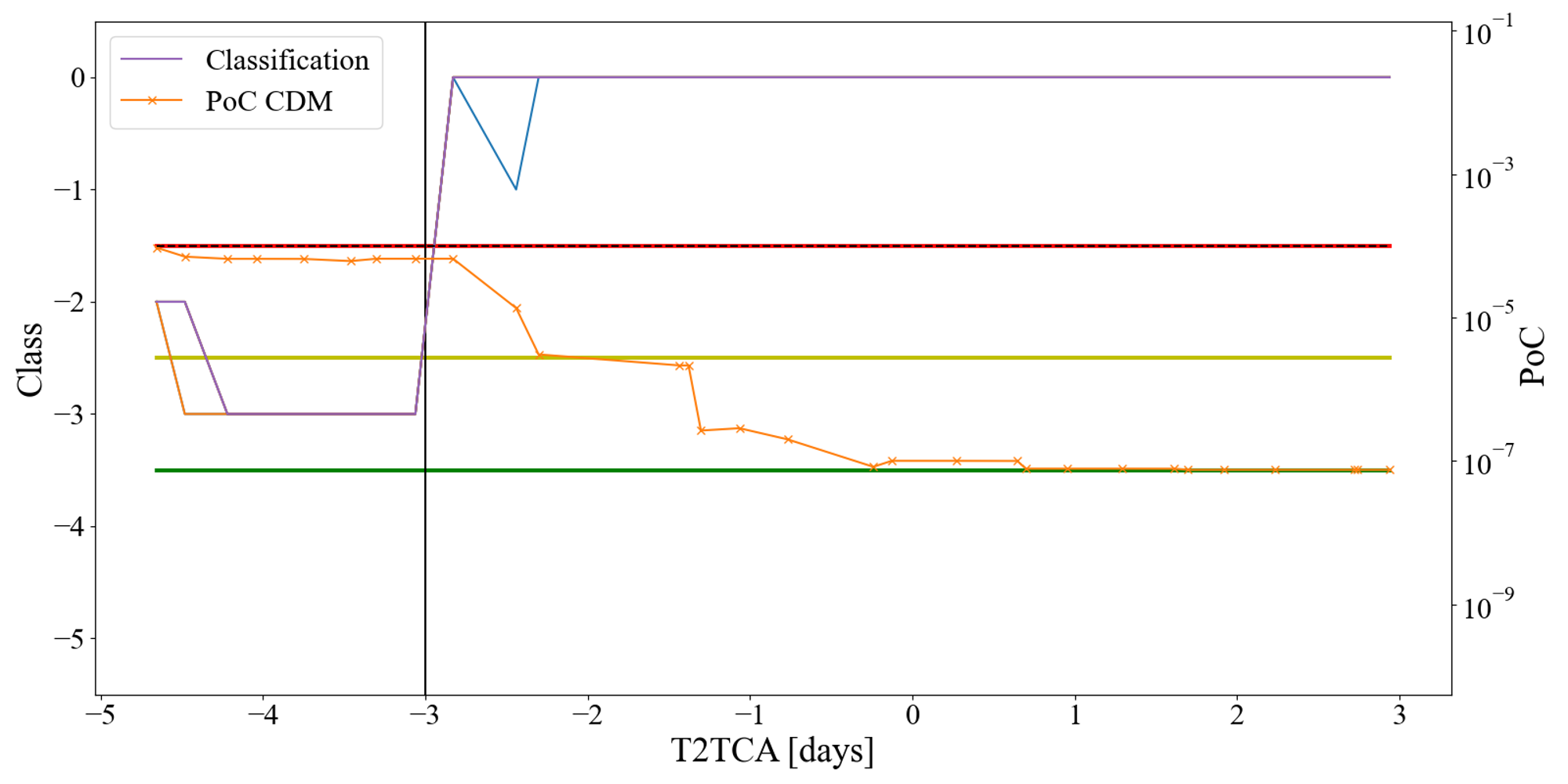}
        \captionsetup{width=0.8\linewidth,format=plain, font=small, labelfont=bf,justification=justified}
        \caption{Collision risk assessment for Event \#3: Uncertain event. Solid narrow lines: evidence-based classification with different number of $\alpha$-cuts: $\#\alpha\text{-cuts}=\left\{1,2,3,4,5,7\right\}$ (note that they overlap each other, so only $\#\alpha\text{-cuts}=7$ is visible for all $t2TCA$ in solid purple; $\#\alpha\text{-cuts}=1$ in solid blue and $\#\alpha\text{-cuts}=2$ in solid orange are visible at one $t2TCA$ each). Crossed-solid line: PoC in the CDMs used by SDO for assessment. Horizontal thick lines: evidence approach safety bands: green, low risk-uncertain boundary; yellow, uncertain-high risk boundary; red, mid term high risk-long term high risk boundary. Dashed black line: Risk threshold (overlapping evidence-based high-risk boundary). Vertical black line: decision time threshold $T_1$. Vertical dashed grey line: TCA.} 
        \label{fig:Clas_NC2_Sc_3_seq}
    \end{figure}

    \cref{fig:Clas_NC2_Sc_3_seq} shows the result of the classification: the event starts at \textit{Class 2}, given the potential high risk suggested by the initial \acp{CDM} but quickly drops to \textit{Class 3} ($t2TCA>T_1$) because of the level of uncertainty and is finally classified as \textit{Class 0 }(for $t2TCA\leq T_1$). In this case, our approach would suggest a further analysis due to the non-zero plausibility of a high \ac{PoC} and a high difference between \ac{Pl} and \ac{Bel}, while the decision made by the \ac{SDO} was to take no further action. The more prudent recommendation coming from our classification system would lead to a further inspection of the \ac{Pl} curve with the realisation that the supporting evidence is small, albeit not zero. 

    \subsubsection{CNES Conjunction Risk Assessment}
    \label{sec:CNES_example}       
   
    In order to compensate for the possible lack of realism of the covariance matrix at \ac{TCA}, \ac{CNES} re-scales both the covariance matrix of the primary and secondary body with two factors, respectively $k_p\in K_P$ and $k_s\in K_S$.
    A scaled \ac{PoC}, called \ac{sPoC}, is obtained by solving the following \ac{PoC} maximisation problem (see \cite{Laporte2014_a,Laporte2014_b}):
    \begin{equation}\label{eq:sPoC}
        \left\{\begin{array}{l}
        sPoC = \max_{k_p\in K_P, k_s\in K_S}PoC(\Sigma)\\
        \text{s.t.} \hspace{10pt}
        \mathbf{\Sigma} = k_p^2\mathbf{\Sigma}_p + k_s^2\mathbf{\Sigma}_s
        \end{array}\right. ,
    \end{equation}
    where $\mathbf{\Sigma}_p$ and $\mathbf{\Sigma}_s$ are, respectively, the primary and secondary covariance matrices in a given \ac{CDM} associated to the conjunction event under consideration.
    
    The two sets $K_P$ and $K_S$ are derived, for each sequence of \acp{CDM}, under the assumption that \acp{CDM} are samples drawn from an underlying distribution, and the last \ac{CDM} contains the most reliable estimation of the position of the two objects. Thus, by using the last \acp{CDM} as a reference, it is possible to compute the Mahalanobis distance of all previous \acp{CDM} from the last one. If one assumes that the uncertainty in position is Gaussian, the Mahalanobis distance should follow a $X^2$ distribution with 3 degrees of freedom. By performing a \ac{KS} test between the distribution of the computed Mahalanobis distances and the theoretical one, and setting a desired level of realism, one can define the sets $K_P$ and $K_S$. More details can be found in \cite{Stroe2021}.
    
    \ac{CNES} decision-making is based on both the value of \ac{sPoC} and a number of geometric considerations. Events with values of $sPoC>5\cdot10^{-4}$ are classified as \textit{High-Interest Event}, the more risky classification level (red level). For values of $10^{-4} < sPoC< 5\cdot10^{-4}$, the event is classified as an \textit{Interest Event}, the second level of risk (orange level). If the value of the $sPoC$ is below those thresholds, caution geometric criteria are applied: miss distance below 1 km or radial distance below 200 m. Note that these threshold values are the default ones and may differ from mission to mission. If the \acp{CDM} are received 4-5 days before the encounter or earlier, no alerts are raised independently of the value of \ac{sPoC}, although the event is placed under study if some of the above criteria are violated. For later \acp{CDM}, alerts may be raised according to the value of \ac{sPoC}. Finally, if the high risk continues after the decision time (usually 2 days before the encounter), a final decision is made before the \ac{TCA}.
    
    In the following, we will test our approach on a real close encounter faced by \ac{CNES} and compare our classification against the one of \ac{CNES}.
    
    \subsubsubsection{Event \#4}
    
    This scenario presents a high-risk collision case for a real close encounter where \ac{CNES} had to implement a manoeuvre to reduce the risk. 
    
\begin{figure}[htb!]
        \centering
        \begin{subfigure}[b]{0.49\textwidth}
            \centering
            \includegraphics[width=\textwidth]{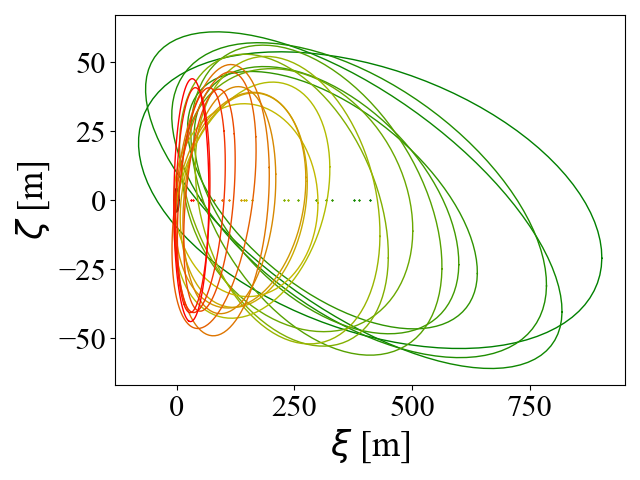}
            \caption{}
            \label{fig:CDM_NC2_Sc_CNES1_a}
        \end{subfigure}
        \begin{subfigure}[b]{0.49\textwidth}
            \centering
            \includegraphics[width=\textwidth]{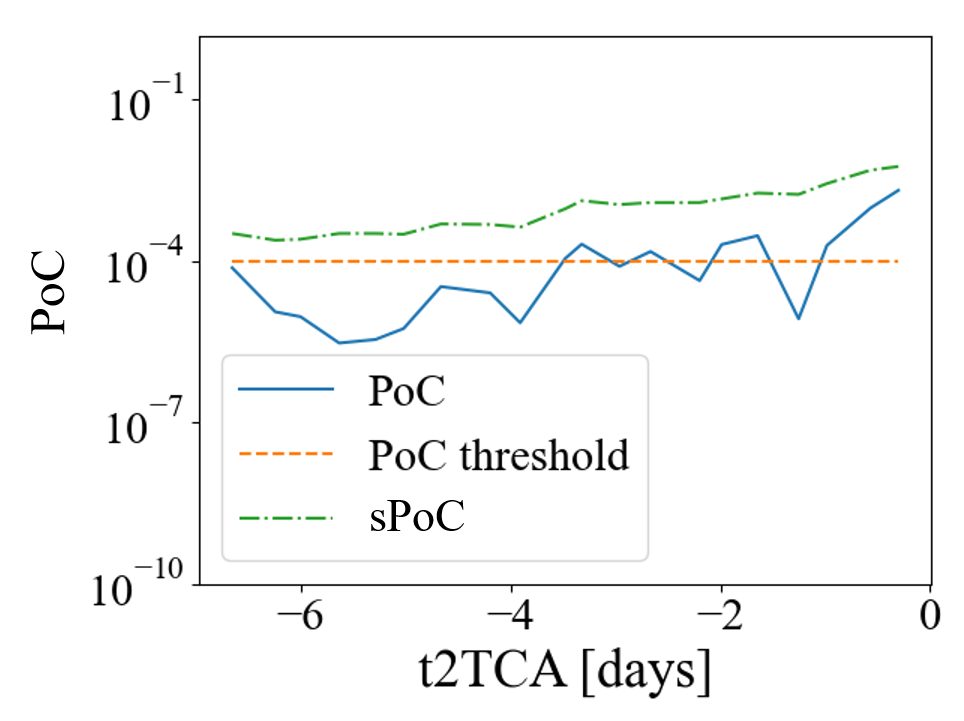}
            \caption{}
            \label{fig:CDM_NC2_Sc_CNES1_b}
        \end{subfigure}
        \captionsetup{width=0.8\linewidth,format=plain, font=small, labelfont=bf,justification=justified}
        \caption{CDM information for example in Event \#4. (a) Uncertain ellipses in the sequence of CDMs. Green ellipses correspond to earlier CDMs, and red ellipses to later CDMs. (b) Evolution of the PoC in the CDMs with the time to the TCA. Blue solid line: PoC; dashed-dotted line: sPoC; orange dashed line: PoC threshold.} 
        \label{fig:CDM_NC2_Sc_CNES1}
    \end{figure}

    \cref{fig:CDM_NC2_Sc_CNES1_a} shows the geometry of the event, where the earlier \acp{CDM} (green ellipses) suggested a low \ac{PoC}, while later \acp{CDM} (red and amber ellipses) suggest a high \ac{PoC}. 
    \cref{fig:CDM_NC2_Sc_CNES1_b} shows the \ac{PoC} and the \ac{sPoC}. The latter is above the threshold $10^{-4}$ from the start and progressively increases while the \ac{PoC} displays a large variability till about a day before TCA. \ac{CNES} classified the event as \textit{High-Interest Event}, meaning that careful monitoring was required, starting from the 12$^{th}$ \ac{CDM} (2.96 days before the \ac{TCA}). The final decision to perform a manoeuvre was taken 30 hours before the encounter. Note that the \ac{CDM} received about a 1.2 days from TCA indicates a \ac{PoC}$<10^{-5}$, well below the risk threshold, while the \ac{sPoC} indicates a risk above $10^{-3}$, which aligns better with the last three \acp{CDM} received between the decision time and the \ac{CAM} execution time).
    
\begin{figure}[htb!]
        \centering
        \begin{subfigure}[b]{0.49\textwidth}
            \centering
            \includegraphics[width=\textwidth]{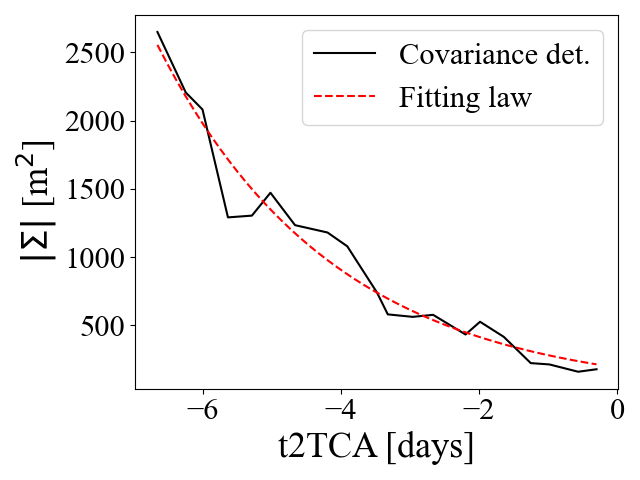}
            \caption{}
            \label{fig:Fitlaw_NC2_Sc_CNES1_a}
        \end{subfigure}
        \begin{subfigure}[b]{0.49\textwidth}
            \centering
            \includegraphics[width=\textwidth]{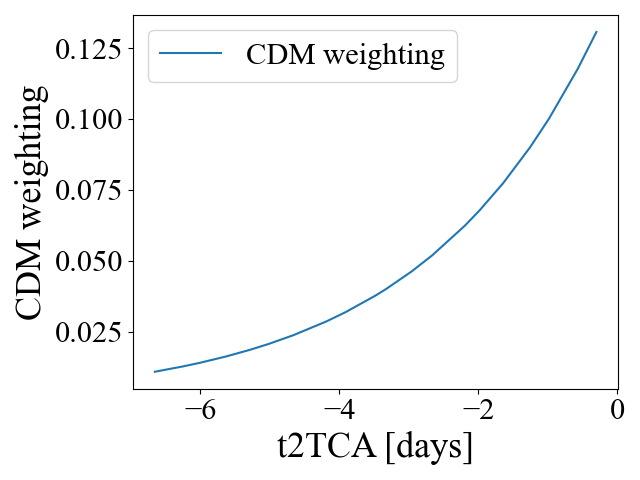}
            \caption{}
            \label{fig:Fitlaw_NC2_Sc_CNES1_b}
        \end{subfigure}
        \captionsetup{width=0.8\linewidth,format=plain, font=small, labelfont=bf,justification=justified}
        \caption{Fitting law to weight the CDMs after having received the whole sequence in Event \#4. (a) Solid black line: value of the determinant from the CDMs, dashed red line: fitting law of the covariance matrix determinant. (b) Weight of the CDMs as a function of the time to the TCA.} 
        \label{fig:Fitlaw_NC2_Sc_CNES1}
    \end{figure}

    The evidence-based analysis was performed following the same approach as for the \ac{SDO} cases, with $\# intervals = \left\{2,3,4,5,6,8\right\}$ intervals per variable and \ac{CDM} weighted according to the exponential law in \cref{fig:Fitlaw_NC2_Sc_CNES1}.
    
\begin{figure}[htb!]
        \centering
        \includegraphics[width=\textwidth]{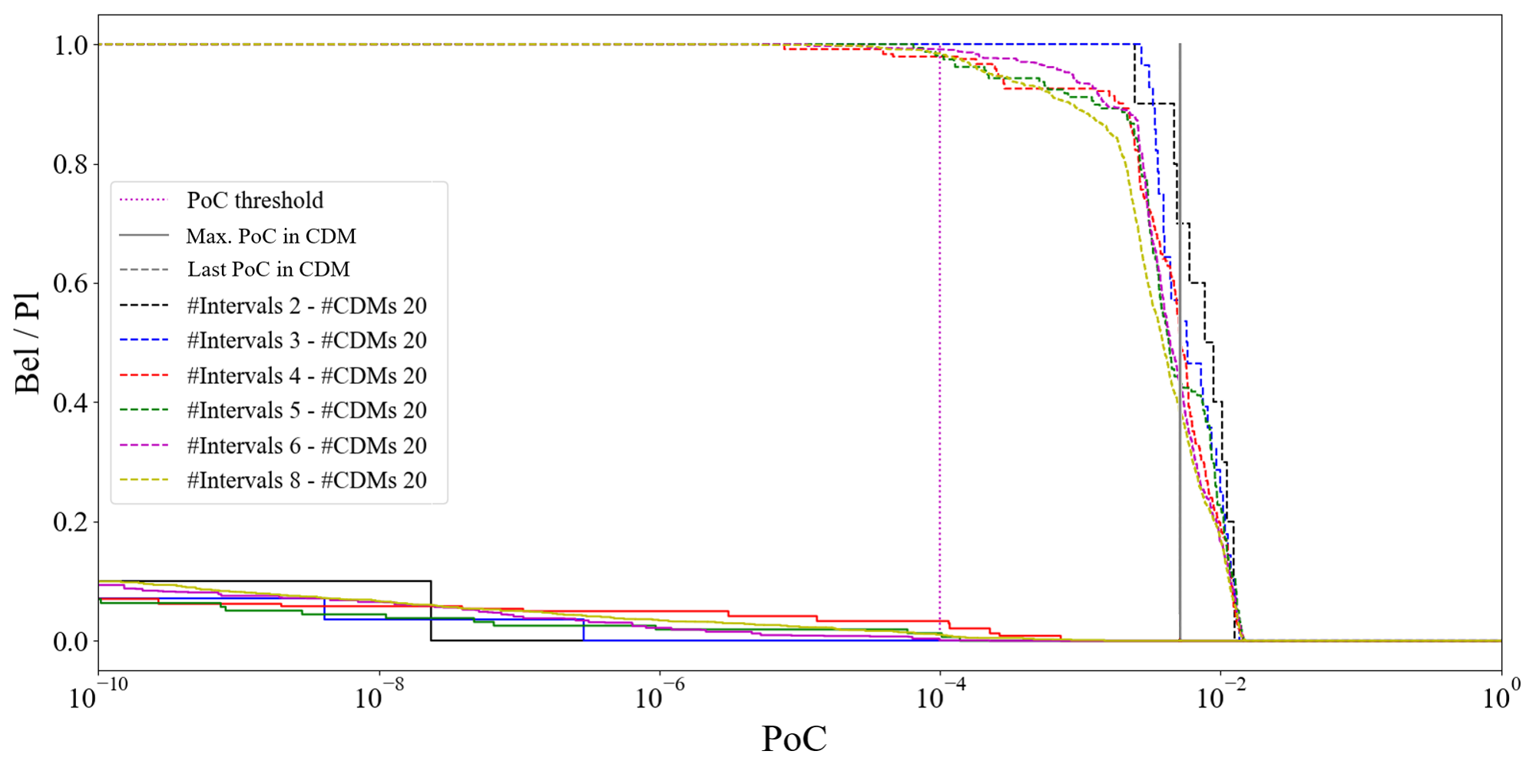}
        \captionsetup{width=0.8\linewidth,format=plain, font=small, labelfont=bf,justification=justified}
        \caption{\textit{Pl} and \textit{Bel} of the PoC after having received the whole sequence of CDMs Event \#4. Solid vertical grey line: maximum PoC in the sequence, dashed vertical grey line: PoC of last CDM, pointed purple line: PoC threshold. For the rest of the colours: Belief in solid lines and Plausibility in dashed lines. Black: 1 $\alpha$-cut per variable (2 intervals per variable, 32 FEs), blue: 2 $\alpha$-cuts, red: 3 $\alpha$-cuts, green: 4 $\alpha$-cuts, purple: 5 $\alpha$-cuts, yellow: 7$\alpha$-cuts.}
        \label{fig:PlBel_NC2_Sc_CNES1}
    \end{figure}

\begin{figure}[htb!]
        \centering
        \includegraphics[width=\textwidth]{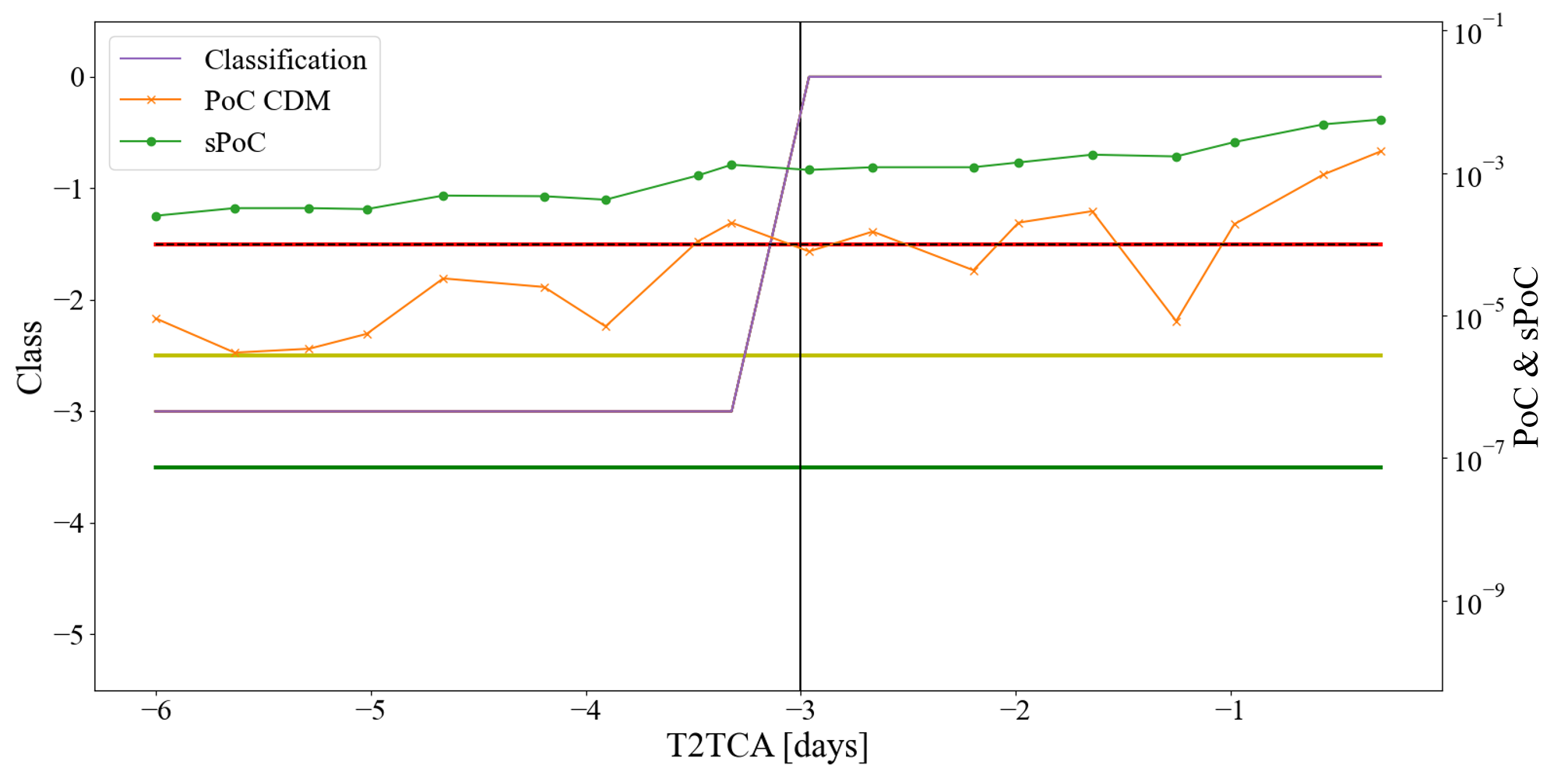}
        \captionsetup{width=0.8\linewidth,format=plain, font=small, labelfont=bf,justification=justified}
        \caption{Collision risk assessment for Event \#4. Solid narrow lines: evidence-based classification with different number of $\alpha$-cuts: $\#\alpha\text{-cuts}=\left\{1,2,3,4,5,7\right\}$ (note that they overlap each other, so only $\#\alpha\text{-cuts}=7$ is visible in solid purple). Crossed-solid line: PoC in the CDMs used by SDO for assessment. Horizontal thick lines: evidence approach safety bands: green, low risk-uncertain boundary; yellow, uncertain-high risk boundary; red, mid-term high risk-long term high risk boundary. Dashed black line: Risk threshold (overlapping evidence-based high-risk boundary). Vertical black line: decision time threshold $T_1$.}
        \label{fig:Clas_NC2_Sc_CNES1_seq}
    \end{figure}

    The \ac{Pl} and \ac{Bel} corresponding to the whole sequence of \ac{CDM} are shown in \cref{fig:PlBel_NC2_Sc_CNES1}, and the classification sequence for different numbers of intervals is shown in \cref{fig:Clas_NC2_Sc_CNES1_seq}. In \cref{fig:PlBel_NC2_Sc_CNES1} one can see that $Pl(PoC_0)$ is nearly 1, and $Pl(sPoC)>0$ along the whole time series. In fact, $Pl=0$ at $PoC\sim10^{-2}$, while $\max(sPoC)=5\cdot10^{-3}$. However, the gap between the \ac{Pl} and \ac{Bel} curves is very high, indicating a degree of uncertainty in the sequence of \acp{CDM}. This is due to the variability in the \acp{CDM}. Thus the event is classified as \textit{Class 0}. 
    
    Although this event is placed in the same class as Event 3, the supporting evidence is quite different. Event 4 has a $Pl\approx 1$ and  \ac{Bel} different from zero at $PoC_0$ while Event 3 has $Bel=0$ and $Pl<0.2$ at $PoC_0$.
    This means that, although in this paper we opted for a very conservative classification of the events such that both Events 3 and 4 fall in the same uncertainty class, a simple analysis of the \ac{Bel} and \ac{Pl} curves would suggests that the available evidence for Event 4 supports a high probability of collision, up to $10^{-2}$ in fact, while for Event 3 the supporting evidence at $PoC_0$ is quite low.
    
    \subsection{Statistical Analysis of CAM Executions}
    \label{sec:num_case_stat}
    
    After having compared the proposed evidence-based conjunction assessment approach against real operations on specific cases, in this section we compare the number \acp{CAM} that our evidence-based approach would recommend over a large number of real conjunctions experienced by a single mission. 
    
    The selected mission is the \ac{ESA} SWARM-A satellite, orbiting in the \ac{LEO} regime (circular polar orbit of 87.7 deg at 511 km of altitude), dedicated to studying the Earth's magnetic field as part of a constellation of three satellites. The mission thresholds to trigger conjunction alerts are $PoC_0=10^{-4}$ and $T_1=72$ hours. Thus, any satellite with a \ac{PoC} above the threshold in the last 3 days would escalate and would require further analysis, and eventually, a possible \ac{CAM} design or execution. Nevertheless, encounters presenting a higher risk or an increasing trend before $T_1$ may be escalated if the operator considers that there is a potential risk for the mission. Finally, the go/no-go decision is subject to operational constraints: the time required to design a \ac{CAM} after receiving the triggering manoeuvre, the possibility to upload and check the design manoeuvre and the ground station availability.
    
    The database of \acp{CDM} includes alerts from 2015 to 2022, with a total of 36,072 events. Overall, most of the events in the database did not represent a threat to the satellite, with only 20 representing escalated events. As explained before, an escalated event is an encounter where the \ac{PoC}, or the \ac{PoC} trend, suggests that the conjunction may be high risk. 
    From those escalated events, only 2 required a \ac{CAM} to be executed.
    
    The evidence-based analysis was performed with the same thresholds as the previous study cases (\cref{table:threshold}): $PoC_0=10^{-4}$, $T_1=3$ days, $T_2=5$ days, $Pl_0=1/243$, $A^*_0 = 0.1$, with $\underline{PoC}=10^{-30}$, and $A_0=3$. The \ac{DKW} bands were obtained assuming a confidence interval of $\delta=0.5$. As shown before, a higher number of $\alpha$-cuts would refine the \ac{Pl} and \ac{Bel} curves, providing closer curves that better represent the actual epistemic uncertainty. However, this is at the expense of increasing the computational cost and with limited impact on the final classification. Thus 2 $\alpha$-cuts (3 intervals) per variable, with a total of 243 \acp{FE} per analysis were used.
    
    Since the evidence-based analysis lacks the real information available in the actual operation of the satellite that may have affected the operator decision (for example, the ground station availability or the mission constraints), the statistics were computed at four decision times: $T_d = 3$ days to the \ac{TCA}, corresponding with the mission time threshold, $T_1$; $T_d = 2$ days to the \ac{TCA}, allowing for more data to arrive; $T_d = 1$ day to the encounter, the usual go-no go decision time in \ac{ESA}'s missions,  \cite{Merz2017}; and the epoch of the last \ac{CDM} in the sequence, $T_d = 0$. For simplicity, we assume that there is no operational constraint that prevents or modifies the final decision and all information is, thus, available.
    
    \cref{table:statistical_SWARMa_} includes the results from the analysis, compared with the actual statistics provided by the \ac{SDO}. It is important to bear in mind the differences between the approaches. An event classified as \textit{Class 3} or \textit{Class 0} (labelled as \textit{Uncertain}), with the evidence-based approach, would not correspond, necessarily, to an escalated event, since the meaning is different: while an escalated event assumes a certain level of risk, a \textit{Class 0}
    or \textit{3}, suggests a degree of uncertainty that requires further investigation before making a final decision. This further investigation might be simply limited to an inspection of the \ac{Bel} and \ac{Pl} curves as in cases 3 and 4 above or might require additional observations. 
    On the other hand, for all \textit{Class 1} events, the recommendation is to perform a \ac{CAM}.

    \begin{table}[hbt!]
        \begin{center}
            %\small
            \captionsetup{width=0.8\linewidth,format=plain, font=small, labelfont=bf, justification=justified}
            \caption{Results from the statistical analysis on the SWARM-A mission, with the SDO approach and the evidence-based approach. Threshold: $PoC_0=10^{-4}$,$T_1=3$ days, $T_2=5$ days, $Pl_0=1/243$. Partition with 2 $\alpha$-cuts per variable. Upper tier: $A^*_0=0.1$ ($A_0=3$); middle tier: $A^*_0=0.5$ ($A_0=15$); lower tier: $A^*_0=0.8$ ($A_0=24$).}
            \begin{tabular}{ l l| l | l l l l l }
                \toprule
                    \multicolumn{2}{c|}{\textbf{SDO}} & \multicolumn{5}{c}{\textbf{Evidence-based}}\\
                \midrule
                     \# events & & $A^*_0$&\# events & $T_d=3$ & $T_d=2$ & $T_d=2$ & $T_d=0$\\
                \midrule
                    Total       & 36,072    & &Total   & 24,296 & 27,918 & 32,108 & 36,072\\
                    Escalated   & 20        & 0.1 & Unc. & 120 & 130 & 172 & 293\\
                    CAM         & 2         & & CAM & 1 & 2 & 3 & 2\\
                \bottomrule
                \midrule
                     &   &  0.5   & Unc. & 102 & 98 & 107 & 154\\
                     &   &     & CAM & 19 & 34 & 68 & 141\\
                \bottomrule
                \midrule
                     &  &  0.8   & Unc. & 95 & 83 & 77 & 75\\
                     &  &     & CAM & 26 & 49 & 98 & 220\\
                \bottomrule
            \end{tabular}
            \label{table:statistical_SWARMa_}
        \end{center}
    \end{table}

    From the upper tier in \cref{table:statistical_SWARMa_} (with $A^*_0=0.1$), one can observe that: i) the total number of events increases with the delay in the decision time because more \ac{CDM}s are available for a decision; ii) the number of manoeuvres proposed by the evidence-based approach is similar to the number of \acp{CAM} proposed by the \ac{SDO} operators; iii)  the evidence-based classification system found many more uncertain cases than the \ac{SDO}.  The Table shows also the number of \acp{CAM} and uncertain events for $A^*_0$ equal to 0.5 and 0.8. As expected, an increase in the values of $A^*_0$ increases the number of \acp{CAM} and reduces the number of uncertain cases.
    
\begin{figure}[htb!]
        \centering
        \begin{subfigure}[b]{0.49\textwidth}
            \centering
            \includegraphics[width=\textwidth]{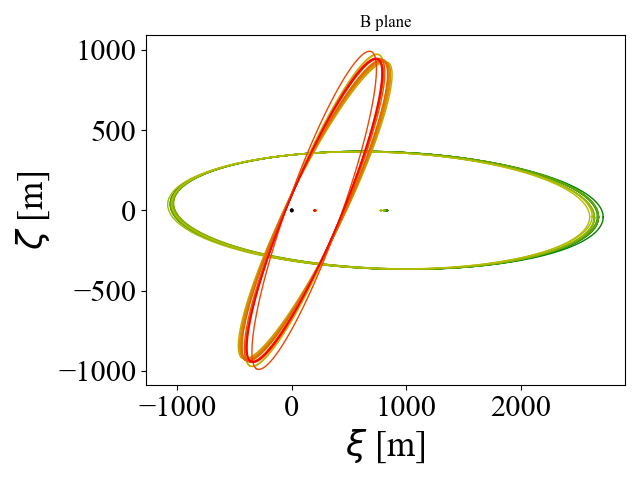}
            \caption{}
            \label{fig:Geometry247116}
        \end{subfigure}
        \begin{subfigure}[b]{0.49\textwidth}
            \centering
            \includegraphics[width=\textwidth]{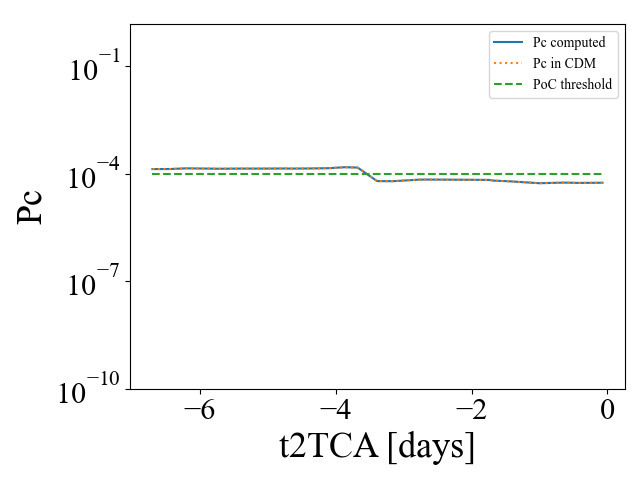}
            \caption{}
            \label{fig:Pc247116}
        \end{subfigure}
        \begin{subfigure}[b]{1.0\textwidth}
            \centering
            \includegraphics[width=\textwidth]{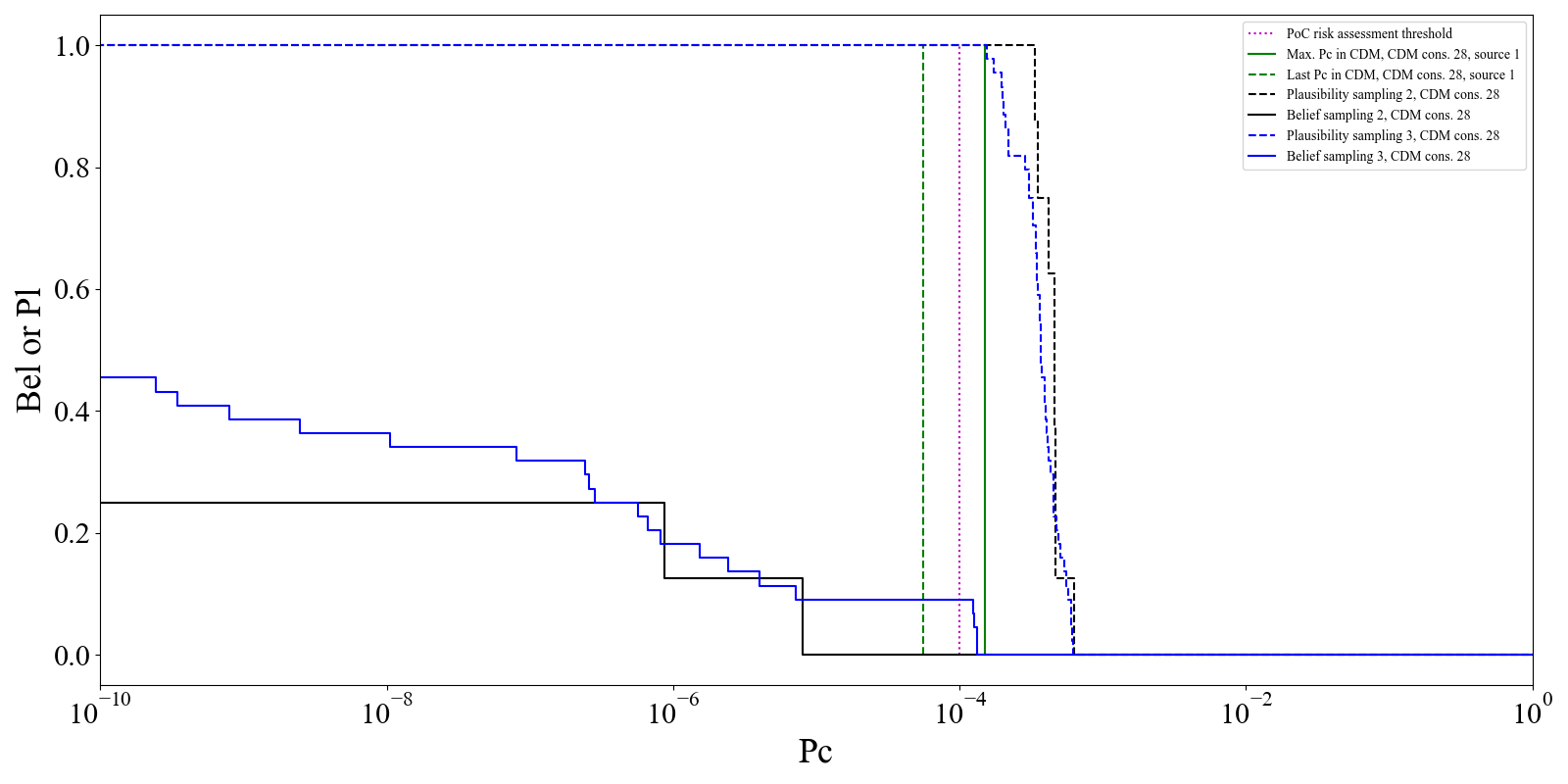}
            \caption{}
            \label{fig:PlBel247116}
        \end{subfigure}        
        \caption{Conjunction event with conflicting CDMs: a) evolution of the relative position distribution on the impact place, b) evolution of the PoC, c) Pl and Bel curves of the whole CDMs sequence.} 
        \label{fig:odd_case}
    \end{figure}

    Even if the $Pl_0$ threshold is quite low, the number of events escalating to \textit{Class 1} remains small. Thus, in this test case, the system is robust enough to remove false negatives without introducing false alerts. Also, the number of \acp{CAM} remains roughly constant independently of the decision time (especially, for the selected default value of $A^*_0=0.1$).
    On the other hand, the number of \textit{Class 0} events is between 6 and 15 times higher than the number of escalated events proposed by \ac{ESOC}. It is here where the evidence-based system differentiates from the probabilistic approach used by \ac{ESOC}. \textit{Class 0} events are those with  $Pl(PoC_0)>Pl_0$, but are still deemed uncertain because $A_{Pl,Bel}>A_0$. $Pl$ captures all realisations, within each Focal Element, that correspond to extreme cases, extreme low or extreme high $PoC$, compatible with the observed sequence of \acp{CDM}. Hence, a large $A_{Pl,Bel}$ with high $Pl$ signifies that there is the evidence that a high $PoC$ event can occur but is uncertain.  
    As in the case of Event 3, many of these cases display a low \ac{Pl} and zero \ac{Bel}. Others present conflicting \ac{CDM}, that cannot be resolved without further observations, or a high \ac{Pl} for high \ac{PoC} values, as in Event 4 but with a low \ac{Bel}.  
    An example can be seen in Figure \ref{fig:odd_case}. The evolution of the combined covariance shows a radical rotation of nearly 90 degrees at -4 days from TCA. The evolution of the \ac{PoC} does not provide any evidence that the covariance had a step change, but remains close to the threshold limit. The evidence-based approach, instead. shows quite some uncertainty and maintains a high $Pl$ till the end of the sequence, suggesting that the event cannot be discarded and requires further analysis.

    Note that the percentage of events in this category increases when delaying the decision. 
    This indicates a growing disagreement among \acp{CDM} in the sequence as the time approaches \ac{TCA}, an aspect usually overlooked by probabilistic-based approaches.

%% file: 5_conclusions.tex
\section{Conclusions}
\label{sec:conclusions}

    This work presented a methodology to model and quantify the epistemic uncertainty in a sequence of \acp{CDM}, and exploit this quantification to make robust decisions about conjunction events. The method was tested against real operations on a number of real scenarios.
    The key working assumption was that the value of the miss distance and covariance matrix in each \ac{CDM} were drawn from a set of unknown distributions. The \ac{DKW} inequality was used to build bounds on this set and derive a set of focal elements, with associated probability mass supporting a given value of the probability of collision. 
    
    The collection of focal elements was used to compute the \ac{Pl} and \ac{Bel} on a given value of the \ac{PoC}. The \ac{Pl} at $PoC_0$, or $Pl_0$ was proposed as a further criterion to make a decision on the actual severity of a conjunction event, while the difference between \ac{Pl} and \ac{Bel}, or $A_{Pl,Bel}$, was proposed as a measure of the uncertainty in the quantification of the \ac{PoC}. 
    
    It was found that when the set of \acp{CDM} contains coherent information over the whole time series, the proposed classification system suggests the same decisions normally made by the \ac{ESA} \ac{SDO}. When the sequence of \acp{CDM} presents a higher degree of variability or a degree of inconsistency the proposed evidence-based approach recommends more conservative decisions compared to the \ac{SDO} but also provides the operator with a quantification of the related uncertainty.
    
    A comparison with the approach used at \ac{CNES}, based on the concept of \ac{sPoC}, showed that the proposed evidence-based approach returns decisions that are less conservative but, at the same time, provides a higher level of information on the uncertainty in the decision. By comparing the \ac{ESA} and \ac{CNES} uncertain cases, it was also found that a further inspection of the \ac{Pl} and \ac{Bel} curves offers a way to disambiguate the events as the different evolution of \ac{PoC} over time is reflected in a lower or higher value of \ac{Pl} and \ac{Bel}.
    
    Finally, a statistical analysis on a database of real encounters of an \ac{ESA} mission showed that the number of recommended \acp{CAM} is similar but the evidence-based approach tends to detect a higher number of uncertain cases that require further analysis. 
    
    Although in our analysis no operational constraints were considered, the number of detected uncertain cases suggests that relying only on the last \ac{CDM} may be too optimistic while the scaled \ac{PoC} approach might be too pessimistic without a further uncertainty quantification. In relation to the uncertain cases, different situations can be found which may lead the operator to take different actions. Further analysis on the treatment of these scenarios should be taken and a threshold tuning analysis using virtual datasets or a mixed dataset of real and virtual \acp{CDM} may help with this task.
    The approach proposed in this paper assumes that no additional information on the CDMs is available nor that information on the uncertainty in the propagation model or individual observations can be used. However, if additional information was available one could improve the quantification of uncertainty of each CDMs and build better defined p-boxes with tighter bounds.
    
    Future work will need to consider the correlation and interdependence among variables during the construction of the focal elements and build a more refined model. Furthermore, the current databases of real \acp{CDM} do not represent a controlled set of events, because the actual outcome is unknown. A representative synthetic database would greatly help in improving the classification system. Last but not least, machine learning can be used to directly classify events from the time series of \acp{CDM}. This approach represents an extension of what was already proposed by the authors and would improve on current efforts to predict the last \ac{CDM} with machine learning as it would embed a quantification of uncertainty in the prediction.

%% file: 9992_acknowledgements.tex
\section*{Acknowledgements}
\label{sec:acknowledgment}

This work was funded by the European Space Agency, through the Open Space Innovation Platform (OSIP), "Idea I-2019-01650: Artificial Intelligence for Space Traffic Management".

The authors would like to thank CNES-Toulouse and the DOA/SME/SE Office for the opportunity to research with them and for sharing really valuable information with us. More specifically, we would like to thank François Laporte for the insightful discussions during the approach development.
The authors would like to thank the ESA's Space Debris Office at ESOC for providing both very useful data and feedback on this work during the research stays at their facilities.